%% file: main.tex
\documentclass[10pt,letter,journal]{IEEEtran}
\input{settings}

\begin{document}
\newcommand{\thetitle}{GNN4REL: Graph Neural Networks for Predicting Circuit Reliability Degradation}

\twocolumn

\title{\thetitle}
\input{authors}
\markboth{IEEE Transactions on Computer-Aided Design of Integrated Circuits and Systems}
{Alrahis \MakeLowercase{\textit{et al.\ }}: GNN4REL: Graph Neural Networks for Predicting Circuit Reliability Degradation}

\IEEEtitleabstractindextext{

\begin{abstract}
\input{abstract.tex}

\end{abstract}

\begin{IEEEkeywords}
Graph neural networks,
Standard-cell libraries,
Static timing analysis,
Transistor aging, 
Reliability estimation
\end{IEEEkeywords}}

\maketitle

\renewcommand{\headrulewidth}{0.0pt}
\thispagestyle{fancy}
\lhead{}
\rhead{}
\chead{\copyright~2022 IEEE.
This is the author's version of the work.
The definitive Version of Record is published in the ESWEEK-TCAD 2022 special issue.}
\cfoot{}

\IEEEdisplaynontitleabstractindextext
\IEEEpeerreviewmaketitle

\setcounter{page}{1}

\input{texfiles/Sec1_Introduction}
\input{texfiles/Sec_background}

\input{texfiles/Sec3_Methodology}
\input{texfiles/Sec_libraries}
\input{texfiles/Sec4_Exp}

\input{texfiles/Sec5_Conclusion}

\section*{Acknowledgment}
This work is supported in part by the Center for Cyber Security (CCS) at New York University Abu Dhabi (NYUAD). Besides, this work is also supported by Advantest as part of the Graduate School ``Intelligent Methods for Test and Reliability'' (GS-IMTR) at the University of Stuttgart. We would like to thank Sami Salamin for his valuable support the RISC-V processor experiments.

\bibliography{main}
\bibliographystyle{IEEEtran}

\input{bios.tex}

\end{document}

%% file: settings.tex
\usepackage{cite}
\usepackage{textcomp}
\usepackage{comment}
\usepackage{verbatim}
\usepackage{stmaryrd}
\usepackage{url}
\usepackage{bm}
\usepackage{pifont}
\usepackage{graphicx}
\usepackage{amsmath,amsfonts,amssymb,amsthm}
\usepackage{subcaption}
\usepackage{setspace}

\usepackage{booktabs} 

\usepackage{pifont}

\usepackage[table,xcdraw]{xcolor}
\usepackage{fancyhdr}
\usepackage{rotating}
\usepackage{graphicx}
\usepackage{verbatim}
\usepackage{hhline}
\usepackage[us]{datetime}
\usepackage{paralist}
\usepackage{color}
\usepackage{soul}
\usepackage{hyperref}

\usepackage[printonlyused]{acronym}
\usepackage{stfloats}

\usepackage{algorithm}
\usepackage{algorithmicx}
\usepackage{algpseudocode}
\usepackage{mathtools}
\usepackage[inline]{enumitem} 
\usepackage{pgfplots}

\usepackage{amsfonts,amsmath}      
\usepackage{nicefrac}       
\usepackage{microtype}      
\usepackage{multirow}
\usepackage{amsthm,bbm}
\usepackage{ctable}
\usepackage{xstring}
\usepackage{tikz}

\DeclarePairedDelimiterX\set[1]\lbrace\rbrace{#1}
\newcommand{\drop}[1]{\textcolor{red}{#1}}
\renewcommand{\drop}[1]{}
\definecolor{cadmiumgreen}{rgb}{0.0, 0.42, 0.24}

\newcommand{\R}{\mathbb{R}}
\newcommand{\Z}{\mathbb{Z}}

\newcommand{\readout}{{\fontfamily{lmtt}\selectfont READOUT}}
\newcommand{\pna}{{\fontfamily{lmtt}\selectfont PNA}}
\newcommand{\sample}{{\fontfamily{lmtt}\selectfont SAMPLE}}
\newcommand{\mlp}{{\fontfamily{lmtt}\selectfont MLP}}
\newcommand{\G}[1][]{
    \ifthenelse{\equal{#1}{}}
    {\mathcal{G}}
    {\mathcal{G}}_{#1}}
\newcommand{\V}[1][]{
    \ifthenelse{\equal{#1}{}}
    {\mathcal{V}}
    {\mathcal{V}_{#1}}}
\newcommand{\E}[1][]{
    \ifthenelse{\equal{#1}{}}
    {\mathcal{E}}
    {\mathcal{E}_{#1}}}
\newcommand{\X}[1][]{
    \ifthenelse{\equal{#1}{}}
    {\bm{X}}
    {\bm{X}^{\paren{#1}}}}
\newcommand{\Anorm}[1][]{
    \ifthenelse{\equal{#1}{}}
    {\widetilde{\bm{A}}}
    {\widetilde{\bm{A}}_{#1}}}

\newcommand{\W}[1][]{
    \ifthenelse{\equal{#1}{}}
    {\bm{W}}
    {\bm{W}^{\paren{#1}}}}

\newcommand{\paren}[1]{\left( #1 \right)}

\graphicspath{{figures/}}

\usepackage{soul}
\definecolor{cadmiumgreen}{rgb}{0.0, 0.42, 0.24}

\newcommand{\purple}[1]{\textcolor{black}{#1}}

\newcommand{\blue}[1]{\textcolor{black}{#1}}
\newdimen\arrayruleHwidth
\makeatletter
\def\Hline{\noalign{\ifnum0=`}\fi\hrule \@height \arrayruleHwidth
\futurelet \@tempa\@xhline}
\makeatother
\newcolumntype{P}[1]{>{\centering\arraybackslash}p{#1}}
\makeatletter
\def\blfootnote{\xdef\@thefnmark{}\@footnotetext}
\makeatother

\makeatletter
\newcommand*{\centerfloat}{%
	\parindent \z@
	\leftskip \z@ \@plus 1fil \@minus \textwidth
	\rightskip\leftskip
	\parfillskip \z@skip}
\makeatother

\shortdate
\settimeformat{ampmtime}

\ifCLASSINFOpdf
\else
\fi
\usepackage{fancyhdr}
\usepackage{amsfonts}       
\usepackage{amsmath}
\usepackage{amssymb}

\usepackage{siunitx}

\usepackage{tikz}
\usepackage{circuitikz}
\usetikzlibrary{calc, fit, positioning}
\usetikzlibrary{shapes, shadows}

\tikzset{%
  cascaded/.style = {%
    general shadow = {%
      shadow scale = 1,
      shadow xshift = -1ex,
      shadow yshift = 1ex,
      draw,
      fill = white},
    general shadow = {%
      shadow scale = 1,
      shadow xshift = -0.5ex,
      shadow yshift = 0.5ex,
      draw,
      fill = white},
    draw,
    fill = white,}
}

\usepackage{pgfplots}
\pgfplotsset{compat=1.15}
\usepgfplotslibrary{colorbrewer}
\usepgfplotslibrary{statistics}
\usepgfplotslibrary{groupplots}

\newcommand{\gnn}{{\fontfamily{lmtt}\selectfont GNN}}

\usepackage{circledsteps}
\pgfkeys{/csteps/inner color=white}
\pgfkeys{/csteps/outer color=none}
\pgfkeys{/csteps/fill color=black}

%% file: authors.tex
\author{
Lilas~Alrahis,~\IEEEmembership{Member,~IEEE,}
Johann~Knechtel,~\IEEEmembership{Member,~IEEE,}
Florian~Klemme,~\IEEEmembership{Member,~IEEE,}
Hussam~Amrouch,~\IEEEmembership{Member,~IEEE,}
and Ozgur~Sinanoglu,~\IEEEmembership{Senior~Member,~IEEE}

\thanks{Manuscript received April 07, 2022; revised June 11, 2022; accepted July 05, 2022. This article was presented in the International Conference on Compilers, Architectures, and Synthesis for Embedded Systems (CASES) 2022 and appears as part of the ESWEEK-TCAD special issue.
}
\IEEEcompsocitemizethanks{
\IEEEcompsocthanksitem Lilas~Alrahis, Johann~Knechtel and Ozgur~Sinanoglu are with the Division of Engineering, New York University Abu Dhabi, Abu Dhabi 129188, UAE (e-mail: lma387@nyu.edu; johann@nyu.edu; ozgursin@nyu.edu).\protect
\IEEEcompsocthanksitem Florian~Klemme and Hussam~Amrouch are with the Department of Computer Science, University of Stuttgart, Stuttgart 70174, Germany (e-mail: {klemme, amrouch}@iti.uni-stuttgart.de).\protect
}

}

%% file: abstract.tex
Process variations and device aging impose profound challenges for circuit designers. Without a precise understanding of the impact of variations on the delay of circuit paths, guardbands, which keep timing violations at bay, cannot be correctly estimated. This problem is exacerbated for advanced technology nodes, where transistor dimensions reach atomic levels and established margins are severely constrained. Hence, traditional worst-case analysis becomes impractical, resulting in intolerable performance overheads. Contrarily, process-variation/aging-aware static timing analysis (STA) equips designers with accurate statistical delay distributions. Timing guardbands that are small, yet sufficient, can then be effectively estimated. However, such analysis is costly as it requires intensive Monte-Carlo simulations. Further, it necessitates access to confidential physics-based aging models to generate the standard-cell libraries required for STA. 

In this work, we employ graph neural networks (GNNs) to accurately estimate the impact of process variations and device aging on the delay of any path within a circuit. Our proposed GNN4REL framework empowers designers to perform rapid and accurate reliability estimations without accessing transistor models, standard-cell libraries, or even STA; these components are all incorporated into the GNN model via training by
the foundry. Specifically, GNN4REL is trained on a FinFET technology model that is calibrated against industrial $14nm$ measurement data. Through our extensive experiments on EPFL and ITC-99 benchmarks, as well as {RISC-V} processors, we successfully estimate delay degradations of all paths -- notably within seconds -- with a mean absolute error down to $0.01$ percentage points.

%% file: texfiles/Sec1_Introduction.tex
\section{Introduction}
\label{sec:Introduction}
The rapid semiconductor-technology scaling has been a primary driver for the industry and an
essential factor for making high-performance electronics widely available. However, technology scaling imposes
enormous challenges when it comes to ensuring the reliability of integrated circuits (ICs) over the projected device lifetime.
Advanced technology nodes show significant increase in the manufacturing process
variability (process variation), making it challenging to predict, for any given IC, the
\textit{timing guardband} that is required to protect the design against timing
violations~\cite{bib:kuhn2011process}.\footnote{%
A timing guardband acts as \textit{safety margin} to prevent 
timing violations. It imposes an additional slack on top of the critical
path delay. Hence, the circuit is clocked at a lower frequency leading to performance ``loss''. Note that guardbands are required, thus the related impact on performance should not be considered a loss a
priori. Rather the value of the guardband is important here: one wants a margin as small as possible, yet
sufficient. Obtaining such optimized value is a challenge especially under the presence of variations.} Furthermore,
device aging (runtime variation), which significantly degrades circuit lifetime and performance, becomes
more dominant in nanoscale nodes.
Without accounting for the impact of variation on the delay of circuit \textit{timing paths},\footnote{%
A timing path is a path between a start point (i.e., a netlist's input port or the clock pin of a sequential element) and an end point (i.e., an output
port or a  data input pin of a sequential element).}
designers cannot guarantee a reliable circuit operation over the
desired lifespan~\cite{bib:amrouch2016reliability}.

\begin{table}[!t]
    \centering
     \captionsetup{justification=centering}
\caption{\purple{Comparison of guardband estimation methods}}
\label{tab:compare}
    \resizebox{\columnwidth}{!}{%

   {\color{black} \begin{tabular}{ccccccc}
        \toprule
        & \multicolumn{3}{c}{Requirements} & & \multicolumn{2}{c}{Capabilities} \\ \cmidrule{2-4} \cmidrule{6-7}
        Method & \multirow{2}{*}{Cell Libraries} & Transistor & Timing & & Process & \multirow{2}{*}{Aging} \\ 
        & & Models & Analysis & & Variation & \\ \midrule
        Conventional STA & One\(^1\) & No & Yes & & No & No\\
        Statistical STA & One (LVF)\(^2\) & No & Yes & & Yes & No \\
        Monte-Carlo STA~\cite{bib:klemme2021machine} & Many & Yes\(^3\)
        & Yes & & Yes & Yes \\
        \textbf{GNN4REL} & No & No & No & & Yes & Yes \\
        \bottomrule
    \end{tabular}}
    }\\[1mm]
\footnotesize
	{\purple{\(^1\)For instance, worst-case corner library. \(^2\)Liberty variation format.}\\
		\purple{\(^3\)To generate the required libraries under variation.}}
\end{table}

\subsection{State-of-the-Art (SOTA) and Their Limitations}
\label{sec:limitations}
Next we outline some general limitations for
existing pre-silicon methods for guardband estimation (See Table~\ref{tab:compare}).

\textbf{Performance Overheads:}
In the worst-case scenario, designers may add pessimistic guardbands (i.e., worst-case margins), which results in
excessive performance overheads and does not allow the circuit to operate at its full potential.

\textbf{Lack of Consideration for Variations in STA:} Static timing analysis (STA) serves to obtain the longest path within a circuit, in terms of signal propagation delay, considering a constant delay value per cell. Such a \emph{critical path} defines the maximum clock speed the circuit can be operated at without causing timing violations. However, conventional STA cannot account for variation due to multiple reasons.
First, the impact of variation on the delay of each cell strongly varies, as demonstrated in \figurename{}~\ref{fig:cell_variabilities}. As a result, variation cannot be sufficiently modeled by simply increasing the critical-path delay by a fixed amount. Second, since the additional delay due to variation will impact each path differently, the critical path needs to be determined with variation in mind. Neglecting variation when analysing the delay of circuit imposes the risk of not correctly identifying the critical path, as the example in \figurename{}~\ref{fig:critical_paths} illustrates.
Since the aforementioned effects are not captured by traditional STA methods, variations cannot be accounted for, and thus, relying on STA standalone leads to estimating overly optimistic timing analysis. Hence, errors due to timing violations can appear during the circuit operation.

\textbf{Computational Cost and Other Limitations of SSTA:}
Statistical STA (SSTA) handles process variation by considering statistical delay distributions rather than constant delay values~\cite{li2013timing,khandelwal2007quadratic}. To enable SSTA, cell libraries (libs) containing variability information are required.\footnote{%
Variability denotes the $\sigma$ of the delay distribution of a given cell divided by $\mu$. The liberty variation format~(LVF) extends the industry-standard liberty format with corresponding $\sigma$ for each contained timing information.}
Hence, SSTA can determine the critical path with respect to its sensitivity to variation. However, SSTA entails significant computational complexity, e.g., Monte-Carlo SPICE simulations to obtain the variability information for the process technology. Although SSTA is a significant step towards better timing guardbands, it still holds some limitations: (i) with
only the average ($\mu$) and standard deviation ($\sigma$) information stored in the lib, SSTA can \textit{only} model Gaussian variability
distributions; {(ii) only the critical path and its variability is examined and then reported, although the actual critical path (in the presence of variability) is likely to change from a circuit
instance to another. This will give a limited/skewed perspective when investigating the actual distribution of
critical-path delays under process variation~\cite{bib:klemme2021machine}}.

\textbf{Need for Standard (Std)-Cell Libs and Transistor Models:} To realize small, yet sufficient, practically
relevant guardbands, note that both STA and SSTA require the generation of
variation- and aging-aware std-cell libs. In other words, only once variation-aware libs are employed, then STA can evaluate the delay of paths under the
	effects that such degradations have.
In turn, this requirement necessitates that designers have
access to confidential transistor models from the foundry (i.e.,
physics-based aging models that well describe the impact of aging on the transistor
parameters). Such access is not always attainable, especially not for advanced nodes.
\begin{figure}[!t]
 \centering \footnotesize
 \input{plots/cell_variability_histogram}
 \caption{Process variation causes different delay variations for different standard-cells, switching transition times, and net capacitance. This distribution is extracted from $1,000$~cell libs, characterized for different instances of process variation. $\mu$ and $\sigma$ denote the
mean and standard deviation, respectively.
 }
 \label{fig:cell_variabilities}
\end{figure}
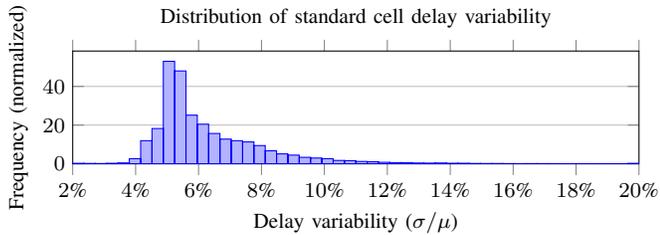
\begin{figure}
 \centering
 \input{plots/critical_path_changes}
 \caption{Process variation and aging impact the location of the critical path. Even in a small circuit such as b01 from ITC-99, the critical path can change rapidly and differently for each chip. The delay analysis here was done using variation-aware libs characterized using SPICE simulations.}
 \label{fig:critical_paths}
\end{figure}
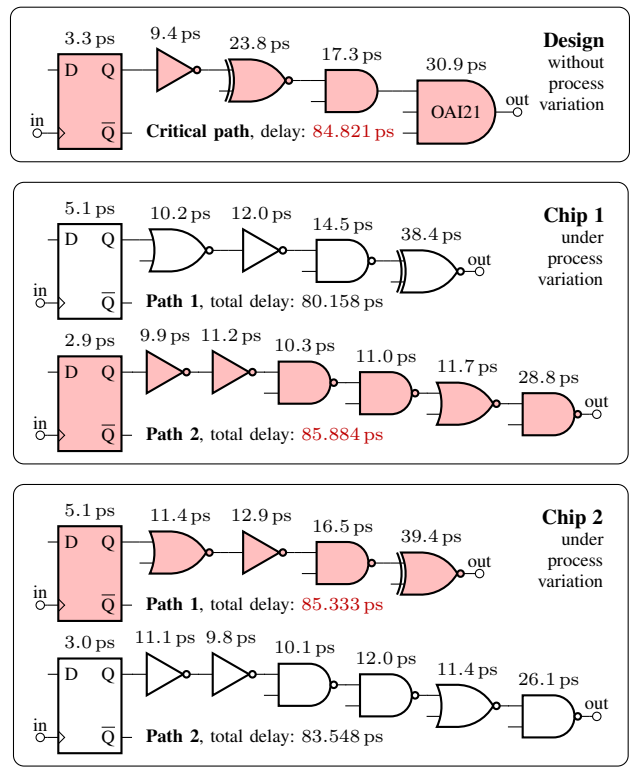
\subsection{Key Research Challenges} 
\label{sec:challenges}

The above review (Sec.~\ref{sec:limitations}) raises an important question: \textit{how to predict process variation- and aging-induced degradation in an efficient manner without relying on std-cell libs, transistor models, or
even STA?} Developing such a versatile reliability assessment framework is an open research problem that
poses the following research challenges.
\begin{enumerate}[leftmargin=*]
\item \textit{Extensive full-circuit analysis:} The delay variation for each cell is impacted, among others,
by its driving cell and load capacitances, i.e., by its location within the netlist. Thus, delay variation is not easy to account for, e.g., neither using a fixed delay increase per cell nor
with a narrow analysis of a \textit{single} critical path. Fig.~\ref{fig:critical_paths} depicts how the critical path of
a circuit can change due to variations, as in featuring a path that was originally critical and then becomes
non-critical whereas another path becomes critical instead.
\item \textit{Handling different types of degradation:}
Depending on the type of considered degradation, different cell libs capturing the corresponding effects need to be generated first. The methods of characterizing process variation and aging-induced degradations are very different, challenging to set up, and expensive in terms of required simulation runtime. Thus, a generic framework that can handle various types of degradation, without requiring the designer to generate specific cell libs beforehand, is desired.
\end{enumerate}
\subsection{Our Novel Concept and Contributions within this Work}
To address the challenges outlined above, we propose GNN4REL, a generic circuit-reliability assessment framework based on graph learning, which is the first of its kind. Graph learning and related graph neural networks (GNNs) are particularly promising here, as this general approach
leverages a naturally matching representation of circuits, unlike other traditional approaches for machine learning.
It is noteworthy that graph learning and GNNs have shown remarkable achievements for other tasks related to circuits, e.g., reverse
engineering of unknown designs~\cite{gnnre} and evaluation of different design-for-trust techniques~\cite{alrahis2021muxlink,alrahis2021untangle,alrahis2021omla,alrahis2021gnnunlockp,9474039}.

\begin{figure}[!t]
\centering
\includegraphics[width=0.45\textwidth]{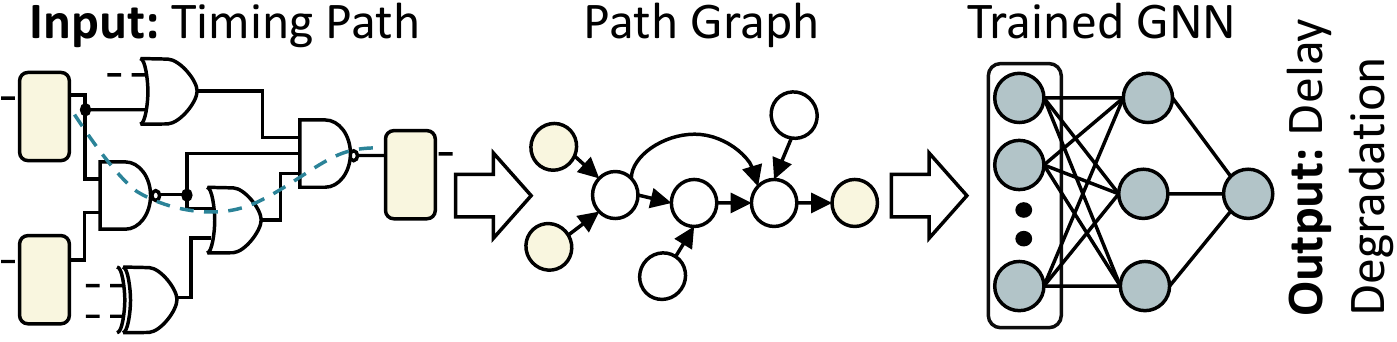}
\caption{Predicting path delay degradation using graph neural networks (GNNs).}
\label{fig:gnn_model_overview}
\end{figure}

\textbf{The key goal} of this work is to train a GNN to predict delay degradations induced by process variation as
well as aging for any given timing path within a circuit. We formulate the problem of estimating delay
degradations as a regression problem and solve the problem using a GNN, as outlined in Fig.~\ref{fig:gnn_model_overview}. This novel framework and its important contributions are supported by the following technical contributions:
\begin{enumerate}[leftmargin=*]
\item We develop a framework for the {path-to-graph conversion (Sec.~\ref{sec:transformation})}, which extracts timing paths from
any gate-level netlist and represents each path as a subgraph. These subgraphs capture the Boolean functionalities of
gates and their directed connectivity within the path. As a result, our platform captures and accounts for the driving cells,
load capacitances, and the position/integration of gates in the netlist when making model predictions.
\item We build a {GNN-based regression model that learns on subgraphs extracted around timing paths (Sec.~\ref{sec3:gnn})}, which automatically
extracts the relevant features of paths that help in estimating the delay degradation. The model can be trained by the
foundry itself, thereby accounting for the manufacturing procedure and aging parameters, and then be shared with the
designers to predict the performance degradations of their circuits. Thus, our platform simplifies the
reliability assessment procedure for the designers (i.e., eliminating the need to build std-cell libs and
run STA), while protecting confidential foundry information.

\end{enumerate}

We demonstrate the effectiveness of GNN4REL on selected ITC-99 and EPFL benchmarks as well as RISC-V
processors. Without loss of generality, our GNN framework is trained on a FinFET technology model that is carefully calibrated against $14nm$
measurement data from Intel obtained from~\cite{bib:natarajan201414nm}.
Note that the calibrations were done for both transistor characteristics as well as variation. Our extensive experiments considering different variation sources and various dataset scenarios show
that GNN4REL achieves excellent prediction performance when predicting delay degradations, reporting a mean absolute error
(MAE) down to $0.01$ percentage points (average of $0.76$). We further release GNN4REL~\cite{GNN4REL_github}.

The remainder of the paper is organized as follows. Sec.~\ref{sec:background} presents the background information, while
Sec.~\ref{sec:methodology} presents the concept and implementation details of GNN4REL. In Sec.~\ref{sec:lib}, the technology calibration and std-cell libs generation (required for
training GNN4REL) are presented. Sec.~\ref{sec:experiments} discusses the experiments and results, \purple{while Sec.~\ref{sec:related_work} covers the related work}. Finally, Sec.~\ref{sec:conclusion} presents the conclusions.

%% file: plots/cell_variability_histogram.tex
\begin{tikzpicture}
    \begin{axis}[
        scale only axis,
        width = 0.85\columnwidth,
        height = 1.5cm,
        title = {Distribution of standard cell delay variability},
        xlabel = {Delay variability (\(\sigma/\mu\))},
        ylabel = {Frequency (normalized)},
        xticklabel={\pgfmathparse{\tick*100}\pgfmathprintnumber{\pgfmathresult}\%},
        xmin = 0.02,
        xmax = 0.20,
        ybar, 
        ymin = 0,
        ymajorgrids,
    ]
    \addplot +[
        hist = {bins = 50, density = true},
    ] table [y=sigmu, col sep=comma] {data/cell_variabilities.csv};
    \end{axis}
\end{tikzpicture}%

%% file: plots/critical_path_changes.tex
\scriptsize 
\begin{tikzpicture}
    \ctikzset{
        logic ports = ieee,
        logic ports/scale = 0.5,
        flipflops/scale = 0.5,
        multipoles/flipflop/font = \scriptsize,
    }
    \node (c0_ff) [flipflop D, fill = red!25] {};
    \draw (c0_ff.pin 3) -- ++(-0.14,0) node (c0_in) [ocirc] {} node (c0_in_text) [above] {in};
    \draw (c0_ff.pin 6) -- ++(0.14,0) node [not port, fill = red!25, anchor = in] (c0_inv1) {}
        (c0_inv1.out) -- ++(0.14,0) node [xnor port, fill = red!25, anchor = in 1] (c0_xnor1) {}
        (c0_xnor1.out) -- ++(0.14,0) node [and port, fill = red!25, anchor = in 1] (c0_and1) {}
        (c0_and1.out) -- ++(0.14,0) node [and port, fill = red!25, number inputs = 3, circuitikz/ieeestd ports/height = 3, anchor = in 1] (c0_oai1) {OAI21}
        (c0_oai1.out) -- ++(0.1,0) node (c0_out) [ocirc] {} node (c0_out_text) [above] {out};
    \node (c0_ff_text) [above = 0cm of c0_ff] {\SI{3.3}{\pico\second}};
    \node (c0_inv1_text) [above = 0cm of c0_inv1] {\SI{9.4}{\pico\second}};
    \node (c0_xnor1_text) [above = 0cm of c0_xnor1] {\SI{23.8}{\pico\second}};
    \node (c0_and1_text) [above = 0cm of c0_and1] {\SI{17.3}{\pico\second}};
    \node (c0_oai1_text) [above = 0cm of c0_oai1] {\SI{30.9}{\pico\second}};
    \node (c0_path_delay) [right = 0.1cm of c0_ff.pin 4] {\textbf{Critical path}, delay: {\color{red!75!black}\SI{84.821}{\pico\second}}};
    \node (c0_boundingbox) [fit =
        (c0_in) (c0_in_text) (c0_ff) (c0_ff_text) (c0_out) (c0_out_text) (c0_path_delay)
        (c0_inv1) (c0_inv1_text)
        (c0_xnor1) (c0_xnor1_text)
        (c0_and1) (c0_and1_text)
        (c0_oai1) (c0_oai1_text)] {};
    \node (c1_ff) [flipflop D, below = 1cm of c0_ff] {};
    \draw (c1_ff.pin 3) -- ++(-0.1,0) node (c1_in) [ocirc] {} node (c1_in_text) [above] {in};
    \draw (c1_ff.pin 6) -- ++(0.1,0) node [nor port, anchor = in 1] (c1_nor1) {}
        (c1_nor1.out) -- ++(0.1,0) node [not port, anchor = in] (c1_inv1) {}
        (c1_inv1.out) -- ++(0.1,0) node [nand port, anchor = in 1] (c1_nand1) {}
        (c1_nand1.out) -- ++(0.1,0) node [xnor port, anchor = in 1] (c1_xnor1) {}
        (c1_xnor1.out) -- ++(0.1,0) node (c1_out) [ocirc] {} node (c1_out_text) [above] {out};
    \node (c1_ff_text) [above = 0cm of c1_ff] {\SI{5.1}{\pico\second}};
    \node (c1_nor1_text) [above = 0cm of c1_nor1] {\SI{10.2}{\pico\second}};
    \node (c1_inv1_text) [above = 0cm of c1_inv1] {\SI{12.0}{\pico\second}};
    \node (c1_nand1_text) [above = 0cm of c1_nand1] {\SI{14.5}{\pico\second}};
    \node (c1_xnor1_text) [above = 0cm of c1_xnor1] {\SI{38.4}{\pico\second}};
    \node (c1_path_delay) [right = 0.1cm of c1_ff.pin 4] {\textbf{Path 1}, total delay: \SI{80.158}{\pico\second}};
    \node (c1_boundingbox) [fit =
        (c1_in) (c1_in_text) (c1_ff) (c1_ff_text) (c1_out) (c1_out_text) (c1_path_delay)
        (c1_nor1) (c1_nor1_text)
        (c1_inv1) (c1_inv1_text)
        (c1_nand1) (c1_nand1_text)
        (c1_xnor1) (c1_xnor1_text)] {};
    \node (c2_ff) [flipflop D, fill = red!25, below = 0.5cm of c1_ff] {};
    \draw (c2_ff.pin 3) -- ++(-0.1,0) node (c2_in) [ocirc] {} node (c2_in_text) [above] {in};
    \draw (c2_ff.pin 6) -- ++(0,0) node [not port, fill = red!25, anchor = in] (c2_inv1) {}
        (c2_inv1.out) -- ++(0,0) node [not port, fill = red!25, anchor = in] (c2_inv2) {}
        (c2_inv2.out) -- ++(0,0) node [nand port, fill = red!25, anchor = in 1] (c2_nand1) {}
        (c2_nand1.out) -- ++(0,0) node [nand port, fill = red!25, anchor = in 1] (c2_nand2) {}
        (c2_nand2.out) -- ++(0,0) node [nor port, fill = red!25, anchor = in 1] (c2_nor1) {}
        (c2_nor1.out) -- ++(0,0) node [nand port, fill = red!25, anchor = in 1] (c2_nand3) {}
        (c2_nand3.out) -- ++(0.1,0) node (c2_out) [ocirc] {} node (c2_out_text) [above] {out};
    \node (c2_ff_text) [above = 0cm of c2_ff] {\SI{2.9}{\pico\second}};
    \node (c2_inv1_text) [above = 0cm of c2_inv1] {\SI{9.9}{\pico\second}};
    \node (c2_inv2_text) [above = 0cm of c2_inv2] {\SI{11.2}{\pico\second}};
    \node (c2_nand1_text) [above = 0cm of c2_nand1] {\SI{10.3}{\pico\second}};
    \node (c2_nand2_text) [above = 0cm of c2_nand2] {\SI{11.0}{\pico\second}};
    \node (c2_nor1_text) [above = 0cm of c2_nor1] {\SI{11.7}{\pico\second}};
    \node (c2_nand3_text) [above = 0cm of c2_nand3] {\SI{28.8}{\pico\second}};
    \node (c2_path_delay) [right = 0.1cm of c2_ff.pin 4] {\textbf{Path 2}, total delay: {\color{red!75!black}\SI{85.884}{\pico\second}}};
    \node (c2_boundingbox) [fit =
        (c2_in) (c2_in_text) (c2_ff) (c2_ff_text) (c2_out) (c2_out_text) (c2_path_delay)
        (c2_inv1) (c2_inv1_text)
        (c2_inv2) (c2_inv2_text)
        (c2_nand1) (c2_nand1_text)
        (c2_nand2) (c2_nand2_text)
        (c2_nor1) (c2_nor1_text)
        (c2_nand3)(c2_nand3_text)] {};
    \node (v1_bb) [draw, rounded corners, fit = (c1_boundingbox) (c2_boundingbox)] {};
    \node (v1_bb_text) [text width = 1.2cm, align = right, anchor = north east] at ($(v1_bb.north east) + (-0.25,-0.25)$) {\textbf{\footnotesize Chip~1} \\ under process variation};
    \node (c3_ff) [flipflop D, fill = red!25, below = 1cm of c2_ff] {};
    \draw (c3_ff.pin 3) -- ++(-0.1,0) node (c3_in) [ocirc] {} node (c3_in_text) [above] {in};
    \draw (c3_ff.pin 6) -- ++(0.1,0) node [nor port, fill = red!25, anchor = in 1] (c3_nor1) {}
        (c3_nor1.out) -- ++(0.1,0) node [not port, fill = red!25, anchor = in] (c3_inv1) {}
        (c3_inv1.out) -- ++(0.1,0) node [nand port, fill = red!25, anchor = in 1] (c3_nand1) {}
        (c3_nand1.out) -- ++(0.1,0) node [xnor port, fill = red!25, anchor = in 1] (c3_xnor1) {}
        (c3_xnor1.out) -- ++(0.1,0) node (c3_out) [ocirc] {} node (c3_out_text) [above] {out};
    \node (c3_ff_text) [above = 0cm of c3_ff] {\SI{5.1}{\pico\second}};
    \node (c3_nor1_text) [above = 0cm of c3_nor1] {\SI{11.4}{\pico\second}};
    \node (c3_inv1_text) [above = 0cm of c3_inv1] {\SI{12.9}{\pico\second}};
    \node (c3_nand1_text) [above = 0cm of c3_nand1] {\SI{16.5}{\pico\second}};
    \node (c3_xnor1_text) [above = 0cm of c3_xnor1] {\SI{39.4}{\pico\second}};
    \node (c3_path_delay) [right = 0.1cm of c3_ff.pin 4] {\textbf{Path 1}, total delay: {\color{red!75!black}\SI{85.333}{\pico\second}}};
    \node (c3_boundingbox) [fit =
        (c3_in) (c3_in_text) (c3_ff) (c3_ff_text) (c3_out) (c3_out_text) (c3_path_delay)
        (c3_nor1) (c3_nor1_text)
        (c3_inv1) (c3_inv1_text)
        (c3_nand1) (c3_nand1_text)
        (c3_xnor1) (c3_xnor1_text)] {};
    \node (c4_ff) [flipflop D, below = 0.5cm of c3_ff] {};
    \draw (c4_ff.pin 3) -- ++(-0.1,0) node (c4_in) [ocirc] {} node (c4_in_text) [above] {in};
    \draw (c4_ff.pin 6) -- ++(0,0) node [not port, anchor = in] (c4_inv1) {}
        (c4_inv1.out) -- ++(0,0) node [not port, anchor = in] (c4_inv2) {}
        (c4_inv2.out) -- ++(0,0) node [nand port, anchor = in 1] (c4_nand1) {}
        (c4_nand1.out) -- ++(0,0) node [nand port, anchor = in 1] (c4_nand2) {}
        (c4_nand2.out) -- ++(0,0) node [nor port, anchor = in 1] (c4_nor1) {}
        (c4_nor1.out) -- ++(0,0) node [nand port, anchor = in 1] (c4_nand3) {}
        (c4_nand3.out) -- ++(0.1,0) node (c4_out) [ocirc] {} node (c4_out_text) [above] {out};
    \node (c4_ff_text) [above = 0cm of c4_ff] {\SI{3.0}{\pico\second}};
    \node (c4_inv1_text) [above = 0cm of c4_inv1] {\SI{11.1}{\pico\second}};
    \node (c4_inv2_text) [above = 0cm of c4_inv2] {\SI{9.8}{\pico\second}};
    \node (c4_nand1_text) [above = 0cm of c4_nand1] {\SI{10.1}{\pico\second}};
    \node (c4_nand2_text) [above = 0cm of c4_nand2] {\SI{12.0}{\pico\second}};
    \node (c4_nor1_text) [above = 0cm of c4_nor1] {\SI{11.4}{\pico\second}};
    \node (c4_nand3_text) [above = 0cm of c4_nand3] {\SI{26.1}{\pico\second}};
    \node (c4_path_delay) [right = 0.1cm of c4_ff.pin 4] {\textbf{Path 2}, total delay: \SI{83.548}{\pico\second}};
    \node (c4_boundingbox) [fit =
        (c4_in) (c4_in_text) (c4_ff) (c4_ff_text) (c4_out) (c4_out_text) (c4_path_delay)
        (c4_inv1) (c4_inv1_text)
        (c4_inv2) (c4_inv2_text)
        (c4_nand1) (c4_nand1_text)
        (c4_nand2) (c4_nand2_text)
        (c4_nor1) (c4_nor1_text)
        (c4_nand3)(c4_nand3_text)] {};
    \node (v2_bb) [draw, rounded corners, fit = (c3_boundingbox) (c4_boundingbox)] {};
    \node (v2_bb_text) [text width = 1.2cm, align = right, anchor = north east] at ($(v2_bb.north east) + (-0.25,-0.25)$) {\textbf{\footnotesize Chip~2} \\ under process variation};
    \node (c0_temp1) at ($(v1_bb_text.east) + (0, 2)$) {};
    \node (c0_temp2) [fit = (c0_temp1)] {};
    \node (v1_bb) [draw, rounded corners, fit = (c0_boundingbox) (c0_temp2)] {};
    \node [text width = 1.2cm, align = right, anchor = north east] at ($(v1_bb.north east) + (-0.25,-0.25)$) {\textbf{\footnotesize Design} \\ without process variation};
\end{tikzpicture}%

%% file: texfiles/Sec_background.tex
\section{Background}
\label{sec:background}
In this section, we present the necessary background information about process variation, device aging, and GNNs. 

\subsection{Process Variation and Device Aging}
\emph{Process variation} occurs due to imperfections in the manufacturing process of the chip. It is a time-independent source of variation that differs for each fabricated chip (and even within the chip itself). Random dopant fluctuation, metal gate granularity, and line-edge roughness are among the typical sources of variation that are considered for state-of-the-art FinFET devices~\cite{bib:zhang2018extraction}. An accurate estimation of process variation is prerequisite to ensure high yield at high performance. If the impact of process variation is underestimated, many fabricated chips will not pass chip testing and the yield is reduced.

In addition to process variation, transistors will also suffer from \emph{aging-induced degradation}. Transistor aging is a time-dependent source of variation that depends on many conditions such as temperature, workload, and projected lifetime. Physical effects that lead to transistor aging include hot-carrier injection (HCI), and importantly, bias temperature instability~(BTI)~\cite{bib:amrouch2016reliability}. BTI is one of the dominant contributors to aging-induced degradations. Over the lifetime of the chip, electrical charges get trapped in the gate oxide of transistors, resulting in increased threshold voltages. In addition, interface traps can be generated at the $Si$-$SiO2$ layer resulting in more undesired charges and hence further increase in the threshold voltage. Consequently, transistor switching times and circuit path delays increase.

We model and include both types of variation (i.e., process variation and aging) in the standard electronic design automation (EDA) tool flows, as later described in Sec.~\ref{sec:creation}.
\subsection{Graph Neural Networks (GNNs)}
The GNN formalism is a dominant paradigm for deep learning with graph structured data. GNNs take a graph as an input and generate an \textit{embedding} (1D vector) for each node in the graph through \textit{neighborhood aggregation/message passing}. The key concept is that the generated embeddings depend on (i) the structure of the graph and (ii) any feature information associated with the nodes. Similar nodes in the graph should be close in terms of distance in the embedding space.

Let $G(V, E)$ be a directed attributed graph; $V$ denotes the set of nodes, and $E$ denotes the set of edges. Each node in the graph $v \in V$ is associated with a feature vector ($\bm{x}_v$) that captures its properties. During neighborhood aggregation ({\textit{aggregate}}), each node receives information (\textit{messages}) from its neighboring nodes $N(v)$, and a new embedding is computed for each node by applying a learnable update function (\textit{update}) on the node's current embedding and the aggregated messages. After $L$ aggregation rounds, each node in the graph is represented by an embedding that captures its properties, the properties of its neighborhood, and its position/integration within the graph. The final node embeddings are then utilized to perform the desired task, such as node classification, graph classification, etc. The neighborhood aggregation procedure is abstracted as follows, where $\bm{z}_v^{(l)}\in \R^f$ indicates the embedding of node $v$ at the $l$-th aggregation round, and $\bm{a}_v^{(l)}$ denotes the aggregated information from $N(v)$ at the $l$-th layer. $Z^{(L)}\in \R^{n \times f}$ indicates the final 2D node embedding matrix, where $n=|V|$.
\begin{equation}
\small
 \bm{a}_v^{(l)} = \textit{aggregate}^{(l)} \left( \left\lbrace \bm{z}_u^{(l-1)} : u \in N(v) \right\rbrace \right) 
\end{equation}
\begin{equation}
\small
 \bm{z}_v^{(l)} = \textit{update}^{(l)} \left( \bm{z}_v^{(l-1)}, \bm{a}_v^{(l)} \right)
\end{equation}

For graph-level tasks, a graph embedding $\bm{y}_G$ that captures the underlying properties of the graph is obtained by applying a \textit{readout} function to the node embeddings. The readout function is typically an order-invariant function such as summing up the node embeddings (i.e., row-wise additions on $Z^{(L)}$).

GNN models mainly differ based on their aggregation and
update functions, with the \textit{mean}, \textit{sum}, and \textit{maximum} aggregation functions being the
most adopted in the state-of-the-art
architectures~\cite{xu2018gin,kipf2016gcn,gilmer2017mpnn,velickovic2019neural}. Different aggregation
functions perform better on different tasks~\cite{xu2018gin}. Therefore, in our work, we employ the
state-of-the-art principal neighborhood aggregation (PNA) model that implements multiple aggregation
strategies instead of a single aggregation function to improve the performance of the GNN model~\cite{corso2020principal}. More details regarding the employed PNA architecture are given in Sec.~\ref{sec3:gnn}.
\begin{figure}[!t]
\centering
\includegraphics[width=0.49\textwidth]{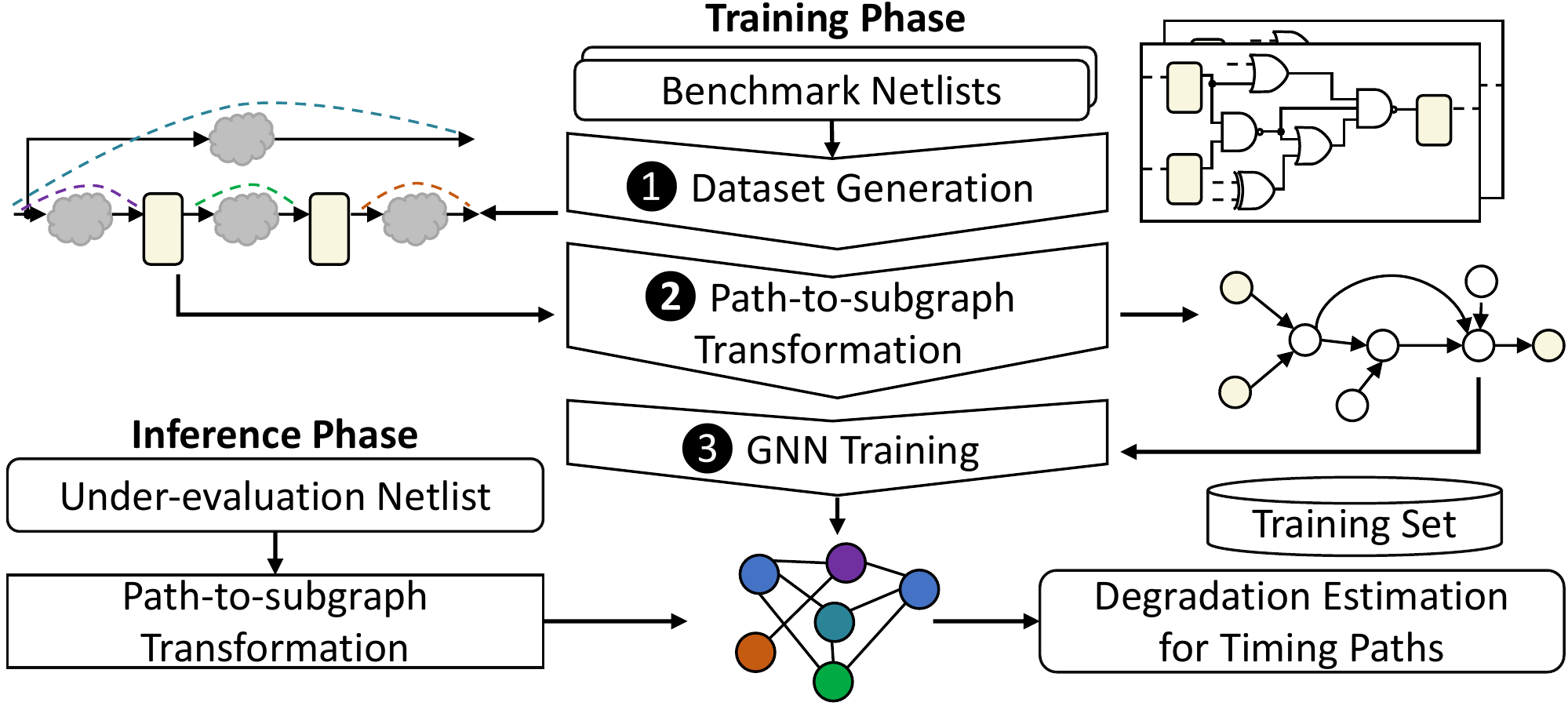}
\caption{The different steps of the proposed GNN4REL framework.}
\label{fig:steps}
\end{figure}

%% file: texfiles/Sec3_Methodology.tex
\section{Our Proposed GNN4REL Framework}
\label{sec:methodology}
 \begin{algorithm}[!t]
 \footnotesize
\caption{Pseudo-code for the proposed GNN4REL platform}
\label{alg:euclid}
\begin{algorithmic}[1] 
\footnotesize
\Require{Netlist $N$, hop-size $h$}
\Ensure{Delay Degradation ($D$)}
\State $P \gets$ PathExtraction($N$)
\State $D\gets \{\emptyset\}$ \Comment{Predicted degradation}
\For{$p\in P$}
\State $S_p\gets(v \in p)$ \Comment{Path target gates}
\State $G_{(S_p,h)}\gets${\sample}($G,h,S_p$) \Comment{Get $G_{(S_p,h)}$}
\State $D.append$({\gnn}($G_{(S_p,h)}$)) \Comment{Get the predictions}
\EndFor
\State \textbf{return} $D$\Comment{Degradation}
\Procedure{{\gnn}}{$G$}
 \State $Z^{(L)} \gets$ {\pna}$(G)$
 \State $\bm{y}_{G} \gets$ {\readout} $(Z^{(L)})$
 \State $\hat{y} \gets$ {\mlp}$(\bm{y}_{G})$
 \State \textbf{return} $\hat{y}$ \Comment{Prediction}
\EndProcedure
\end{algorithmic}
\end{algorithm}
In this section, we provide an overview of the GNN4REL framework (summarized in Fig.~\ref{fig:steps} and Algorithm~\ref{alg:euclid}).

\subsection{Path-to-Subgraph Transformation}
\label{sec:transformation}
We represent each gate-level netlist as a directed graph $G = (V, E)$, where $V$ represents the set of nodes (i.e., gates, primary inputs (PIs), and primary outputs (POs)), while $E$ represents the set of edges (i.e., interconnects). We chose a directed representation to capture the direction of the timing paths (i.e., start point, end point, and order of gates). Each node in the graph $v \in V$ is initialized with a feature vector $\bm{x}_{v}$ that captures its properties (more on that in Sec.~\ref{sec:feature_vector}). $X \in \Z^{n \times k}$ is the 2D matrix containing node features, where $k$ denotes the length of the feature vector.
\subsubsection{Subgraph Extraction}
First, timing paths are extracted from the gate-level netlist and grouped into set $P$ (line 1 in Algorithm~\ref{alg:euclid}). Then, the nodes forming a specific path $p \in P$ are grouped into set $S_p$ (i.e., $\cup \{v \in p\}$) (line 4). We refer to the path nodes as \textit{target nodes}.

Given the netlist graph $G$, an \textit{$h$-hop enclosing subgraph} $G_{(S_p,h)}$ is extracted around the target nodes (line 5). Let $d(i,j)$ represent the shortest path distance between nodes $i$ and $j$, then $G_{(S_p,h)}$ is extracted from $G$ by $\cup_{j\in S_p} \{i~|~ d(i,j) \leq h\}$. We extract an $h$-hop subgraph around the target path to capture the position of the path within the netlist and collect information regarding the driving cell and load capacitance of the gates in the path, which all impacts delay.

\begin{figure}[!t]
\centering
\includegraphics[width=0.49\textwidth]{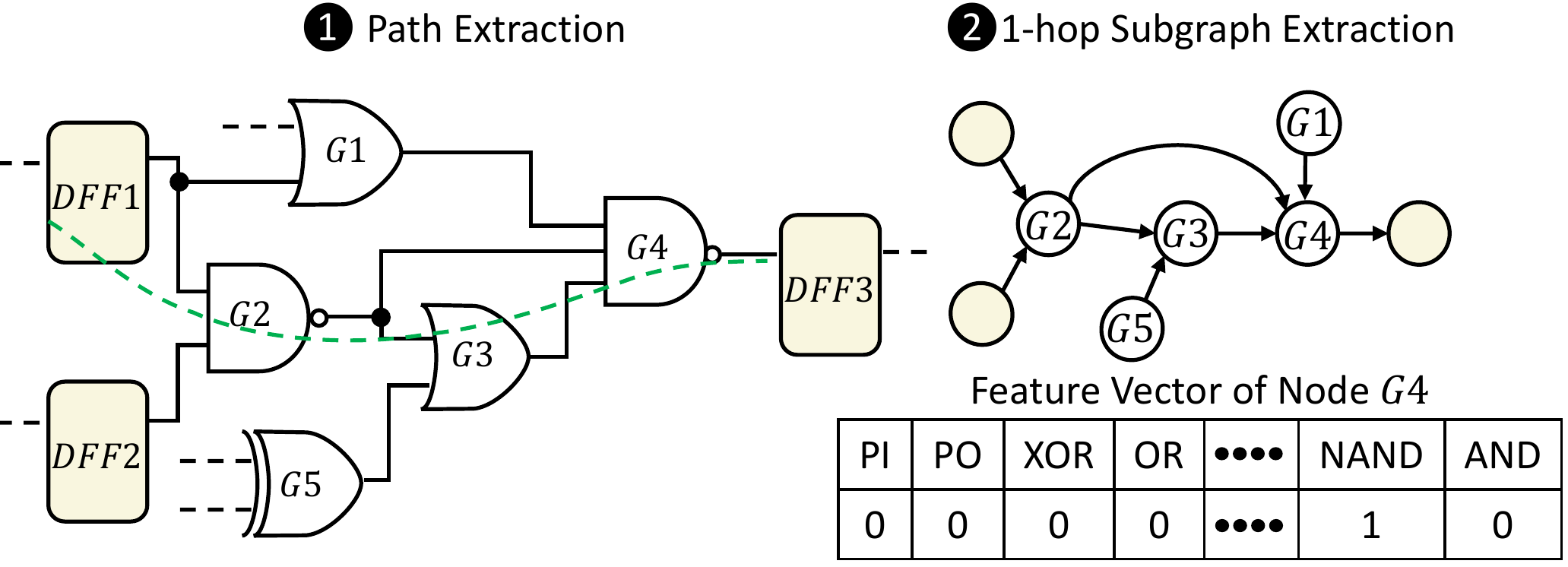}
\caption{Proposed path-to-subgraph transformation.}
\label{fig:conversion}
\end{figure}

An example of $1$-hop subgraph extraction is illustrated in Fig.~\ref{fig:conversion}. The extracted timing path highlighted with a dashed green line includes the gates $\{G2, G3, G4\}$, which form the set $S_p$ (see \Circled{\scriptsize\textbf{1}}). To extract a $1$-hop subgraph around the target path, all the $1$-hop neighbors of the gates in $S_p$ are included in the subgraph, alongside the target gates (see \Circled{\scriptsize\textbf{2}}). The list of $1$-hop neighbors includes $\{DFF1, DFF2, G5, G1\}$.

\subsubsection{Feature Vector}
\label{sec:feature_vector}
$\bm{x}_v$ is a one-hot encoded vector that represents the node's Boolean functionality, or, indicates, if applicable, that it is a PI or a PO. The length of the feature vector $k$ depends on the number of Boolean functions available in the target std-cell lib. The feature vector example in Fig.~\ref{fig:conversion} indicates that node $G4$ is a NAND gate.

\subsection{Employed Graph Neural Network Model}
\label{sec3:gnn}
We employ the state-of-the-art PNA GNN model to perform graph-level regression~\cite{corso2020principal} (line 6 in Algorithm~\ref{alg:euclid}). In this context, a graph represents the extracted subgraph around a target timing path. The PNA employs four statistical aggregators, i.e., $\mu$, \textit{maximum} ($\max$), \textit{minimum} ($\min$), and $\sigma$, so that each node is aware of the distribution of its incoming messages.

The aggregation functions are listed below. $Z^{(l)}$ are the nodes' embeddings at layer $l$. \textit{ReLU} is the rectified linear unit used to avoid negative values caused by numerical errors and $\epsilon$ is a small positive number to ensure $\sigma$ is differentiable.
\begin{equation}
\small
 \label{eq:mean}
 \mu_v(Z^{(l)}) = \frac{1}{| N(v) |} \sum_{u \in N(v)} \bm{z}_u^{(l)}
\end{equation}
\begin{equation}
\small
 \label{eq:max}
 \text{max} _v(Z^{(l)}) = \max_{u \in N(v)} \bm{z}_u^{(l)}
\end{equation}
\begin{equation}
\small
 \text{min} _v(Z^{(l)}) = \min_{u \in N(v)} \bm{z}_u^{(l)}
\end{equation}
\begin{equation}
\small
\sigma_v (Z^{(l)}) = \sqrt{ReLU \left( \mu_v({Z^{(l)}}^2) - \mu_v{(Z^{(l)})}^2 \right) + \epsilon}
\end{equation}

\textit{Degree scalers} allow the network to attenuate or amplify signals based on the degree of each node, i.e., the number of messages being aggregated. PNA uses the logarithmic scaler $S(d, \alpha)$ presented below, where $\alpha$ is a variable parameter that is positive for amplification, negative for attenuation, or zero for no scaling. $\delta$ is a normalization parameter computed over the training set, and $d$ denotes the degree of the target node.
\begin{equation}
\small
\label{eq:Scaler}
S(d, \alpha) = \left(\frac{\log(d + 1)}{\delta} \right)^\alpha, \quad d>0, \quad -1 \leq \alpha \leq 1
\end{equation}

\begin{figure}[!t]
\centering
\includegraphics[width=0.8\columnwidth]{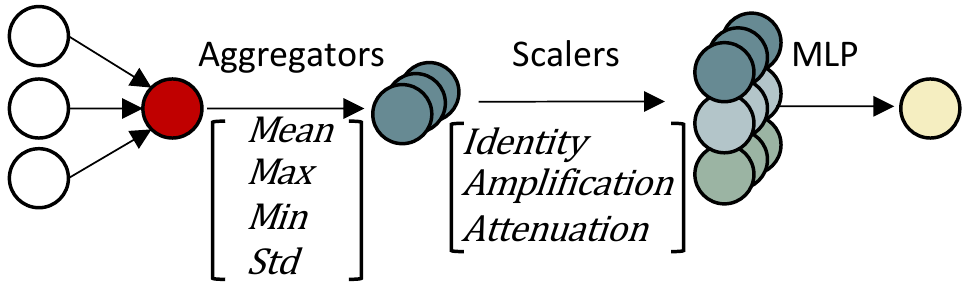}
\caption{The principal neighborhood aggregation (PNA) architecture employs four statical aggregators and three degree scalers~\cite{corso2020principal}.}
\label{fig:pna}
\end{figure}

Note that the PNA model employs a logarithmic scaler instead of a linear scaler since the latter would cause an exponential amplification of both the aggregated information and the gradients after multiple GNN layers. Such an exponential amplification, in turn, would reduce the ability of a GNN to generalize to unseen, possibly larger graphs~\cite{corso2020principal}.

The degree scalers are combined with the aggregator functions as follows, where $\otimes$ represents the tensor product and $I$ represents the identity matrix (i.e., no scaling).
\begin{equation}
\small
\label{eq:PNA}
\bigoplus =
\underbrace{
\begin{bmatrix}
I \\ S(D, \alpha=1) \\
S(D, \alpha=-1)
\end{bmatrix} }_{\text{scalers}}
\otimes
\underbrace{\begin{bmatrix}
\mu \\ \sigma \\ \max \\ \min
\end{bmatrix}}_{\text{aggregators}}
\end{equation}

A PNA layer used in GNN4REL can be abstracted as follows, where $M$ and $U$ are linear layers.

\begin{equation}
\small
\label{eq:PNAmpnn}
\bm{z}_v^{(l+1)} =
U \left( \bm{z}_v^{(l)},
\underset{(u,v) \in E}{\bigoplus}
M \left( \bm{z}_v^{(l)}, \bm{z}_u^{(l)} \right) \right)
\end{equation}

A diagram for the PNA layer is illustrated in Fig.~\ref{fig:pna}, where MLP represent a multi-layer perceptron. After $L$ PNA layers (line 10 in Algorithm~\ref{alg:euclid}), a readout layer is added to obtain a graph-level embedding $\bm{y}_G$ (line 11), which gets passed to an MLP to generate the prediction $\hat{y}$ (line 12). Details regarding the number and dimension of layers are included in Sec.~\ref{sec:exp_gnn}.

\subsection{Dataset Generation}
\label{sec:dataset_generation}
We consider three scenarios of dataset generation to demonstrate the generic nature of our proposed platform. In all scenarios, the setup requires the generation of a dataset from which a training and a validation set are extracted. A dataset contains a list of timing paths extracted from gate-level netlists. The training and validation sets include labeled timing paths (i.e., known delay-degradation percentages), while the testing set includes unlabelled timing paths.

\subsubsection{Self-Referencing}
\label{sec:scenario:self-ref}
The timing paths of a single design are split based on an $81$:$10$:$9$ training:validation:testing ratio. We expect
GNN4REL to achieve the best prediction performance in this setup as it captures the design characteristics
during training (without seeing the exact testing paths).

\subsubsection{Single-Design}
\label{sec:scenario:single-des}
The timing paths of a specific design, referred to as $X$, are used for training and validation based on a $90$:$10$ training:validation ratio, and the timing paths of another design, referred to as $Y$, are used for testing. The goal is to show that GNN4REL can generalize to unseen designs. For example, GNN4REL can be trained on the b14 benchmark and then evaluated on the rest of the ITC-99 benchmarks.

\subsubsection{Design Dataset} 
\label{sec:scenario:des-lib}
We further consider a design dataset (i.e., a collection of gate-level netlists) for training GNN4REL. The design dataset does not include the target design but includes designs with a similar design structure. For example, when predicting the degradation of the b17 benchmark from ITC-99, only the timing paths of b14, b15, b20, b21, and b22 are used for training:validation based on a $90$:$10$ ratio.

%% file: texfiles/Sec_libraries.tex
\section{Technology Calibration \break and Std-Cell Libraries Creation}
\label{sec:creation}

\label{sec:lib}

We utilize mature \SI{14}{\nano\metre} FinFET technology and commercial std-cell characterization tool flows to ensure the generation of accurate training data for realistic circuit timing analysis. This requires careful calibration of the underlying transistor model against silicon measurements, as well as the characterization a wide variety of std-cell libs to capture the technology for different operating conditions. As a result, we generate std-cell libs for the technology under typical conditions, under the impact of process variation, and under the impact of transistor aging as exhibited at the end of the lifetime of a chip.
\subsection{Technology Calibration}

\begin{figure}[!t]
 \centering \footnotesize
 \input{plots/device_calibration_IdVgs}
 \input{plots/device_calibration_IdVds}
 \caption{The industry compact model for FinFET (BSIM-CMG) is carefully calibrated to reproduce Intel \SI{14}{\nano\metre} measurement data extracted from~\cite{bib:natarajan201414nm}. As demonstrated in the plots, SPICE simulations (using our calibrated models) achieve an excellent agreement with the measurement data for both nFinFET and pFinFET devices. The top figure~(a) shows the validation for the case of $I_{ds}$-$V_{gs}$ at high and low $V_{ds}$ biases. \purple{The arrows in the figure indicate which curves belong to which y-axis.} Bottom figure~(b) shows the validation of $I_{ds}$-$V_{ds}$ at various $V_{gs}$ biases. Details on our calibration available in~\cite{calibration_tcas1}.}
 \label{fig:device_calibration}
\end{figure}
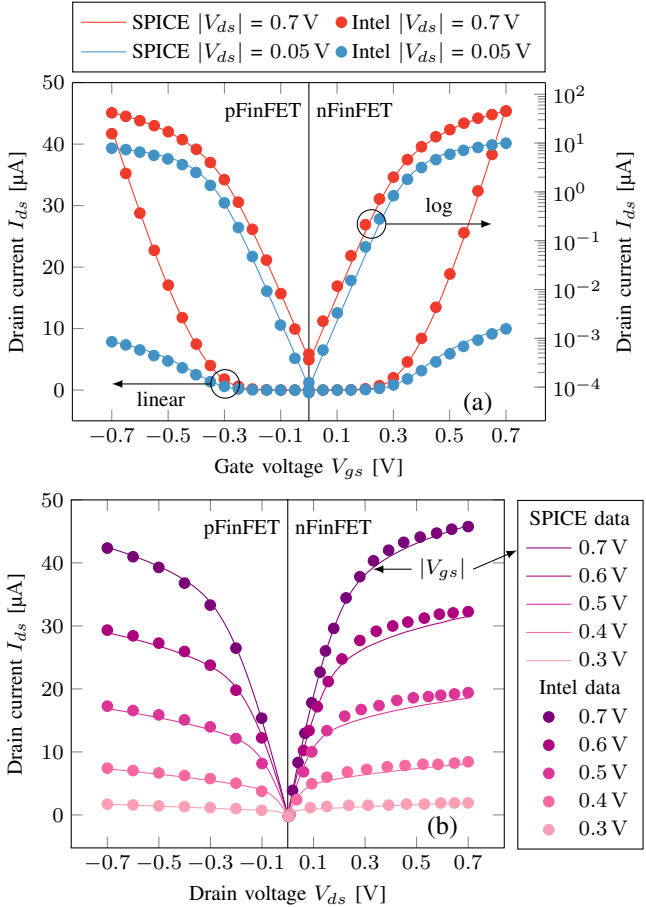

\begin{figure}[!t]
 \centering \footnotesize
 \input{plots/variability_calibration}
 \caption{Variability calibration of our FinFET compact model against Intel \SI{14}{\nano\metre} measurement data~\cite{bib:natarajan201414nm}. The regression line obtained from Monte-Carlo SPICE simulations on the transistor model is in good agreement with the data from the variability measurements. Details on our calibration available in~\cite{calibration_tcas1}.}
 \label{fig:variability_calibration}
\end{figure}
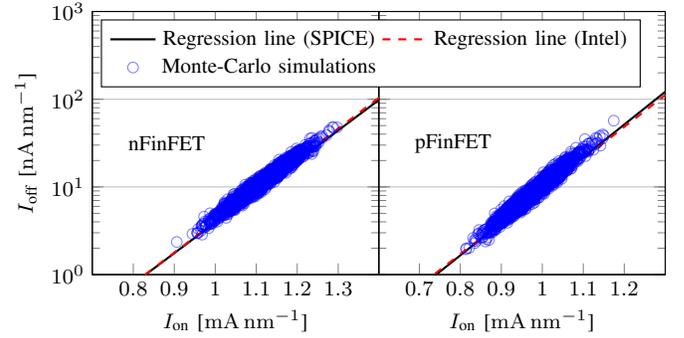

For our analog SPICE simulations, the industry-standard compact model for FinFET technology (BSIM-CMG)~\cite{bib:bsimcmg} is used as underlying transistor model. All parameters of the compact model are carefully calibrated to reproduce measurements of the Intel production-quality \SI{14}{\nano\metre} manufacturing process. Transistor measurements for validation are extracted from~\cite{bib:natarajan201414nm}. \figurename{}~\ref{fig:device_calibration} demonstrates the excellent agreement between SPICE simulation results obtained with our calibrated transistor model and Intel measurement data. As can be seen from the figure, validation is performed for both n-type and p-type FinFET transistors, as well as for multiple \(I_{ds}\)-\(V_{gs}\) and \(I_{ds}\)-\(V_{ds}\) biases, to ensure a holistic representation of the technology and accurate simulation for all required corner cases. In addition, the compact model is further calibrated to reflect the technology under process variation. The considered sources of variation include the gate length~(\(L_g\)), fin thickness~(\(T_\text{fin}\)), fin height~(\(H_\text{fin}\)), \(\text{SiO}_2\) equivalent gate dielectric thickness~(\(EOT\)), and the work-function of the gate~(\(\phi_g\)). \figurename{}~\ref{fig:variability_calibration} demonstrates the variability calibration as an \(I_\text{on}\)-\(I_\text{off}\) plot with regression lines obtained from Monte-Carlo SPICE simulations. The Intel reference regression line is once again extracted from~\cite{bib:natarajan201414nm}.
As shown, our calibrated variability parameters are in excellent agreement with measurements of \SI{14}{\nano\metre} FinFET variability. 

\subsection{Std-Cell Library Generation}

\begin{figure}[!t]
\centering \footnotesize
\includegraphics[width=0.49\textwidth]{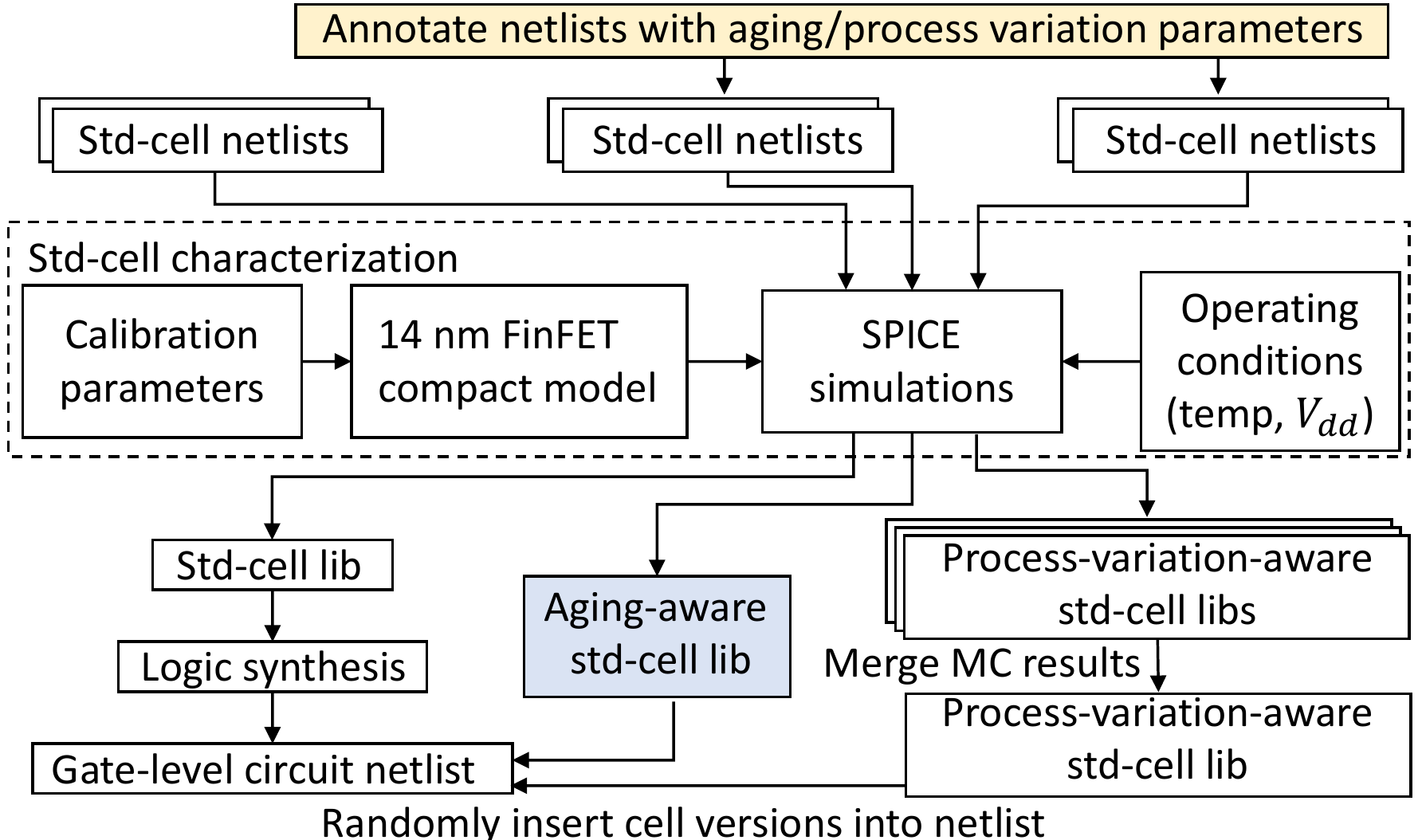}
\caption{Cell library generation workflow for process variation and aging.}
\label{fig:cell_library_generation}
\end{figure}

With an accurately calibrated transistor model, we can characterize a full std-cell lib using accurate SPICE simulations. To this end, the commercial \textit{HSPICE} from \textit{Synopsys} tool flows was used. The required std-cell netlists are obtained from the NanGate \SI{15}{\nano\metre} open-source cell lib~\cite{bib:silvaconangate}. All std-cell netlists are also annotated with post-layout parasitic resistances and capacitances. A commercial characterization tool flow~\cite{bib:primelib} is then employed to instruct extensive SPICE simulations, determining all characteristics of the std-cells, including signal propagation delays, transition times, pin capacitances, switching power, and leakage power. The resulting std-cell lib is stored in the non-linear delay model~(NLDM)-format, in which each data point is represented by a \(7 \times 7\)-matrix to account for the different input signal slews and output load capacitances an std-cell can experience, depending on its position in the circuit. Without any further adjustments to the transistor model or cell netlists, the resulting cell lib reflects the manufacturing process under typical conditions and is suitable for circuit synthesis.

To generate std-cells for different instances of process variation, the transistor model and the std-cell netlists are prepared to accept parameter overwrites for each individual transistor instance. Afterward, each transistor in each cell netlist is annotated with a random set of variability parameters (\(L_g\), \(T_\text{fin}\), \(H_\text{fin}\), \(EOT\), \(\phi_g\)) following the expected distribution of the parameters under process variation. With the annotated cell netlists, std-cell lib characterization is performed as usual. The obtained cell lib reflects each std-cell under the impact of one instance of process variation. Afterward, the aforementioned steps are repeated in a Monte-Carlo fashion, applying new random variability parameters to the std-cell netlists in each iteration. The resulting set of cell libs form a collection of std-cells, characterized under different instances of process variation. With such a collection, 
accurate STA for entire circuits under process variation can be achieved.

The gate-level netlist of a circuit is obtained by performing logic synthesis with the baseline std-cell lib (i.e.,~the reference lib in which neither process variation nor transistor aging is applied). Afterward, each cell instance in the gate-level netlists is replaced by a random version of that cell under process variation. 
All characterized cells under process variation are suffixed with an individual index and merged into one large cell lib. The entire workflow is also outlined in \figurename{}~\ref{fig:cell_library_generation}.

For aging-aware std-cell libs, the approach is very similar to the above. However, instead of the variability parameters, the threshold voltage ($V_{th}$) of each transistor is increased, to reflect the impact of aging. In the industry-standard compact model for FinFET transistors (BSIM-CMG), the \emph{dvtshift} parameter is used to model the major impact of aging. In addition, we also adjust the \(C_{it}\) parameter to control capacitance change due to interface traps. This is, in fact, needed to also reflect the aging-induced degradation in the sub-threshold swing ($SS$) of the transistor. With all parameters annotated in the cell netlists, cell lib characterization is performed to generate the corresponding aging-aware cell libs. These cells can be inserted into gate-level netlists to enable accurate STA for circuits under the impact of aging. 


%% file: plots/device_calibration_IdVgs.tex
\begin{tikzpicture}
    \pgfplotsset{
        set layers = standard,
        scale only axis,
        width = 0.71\columnwidth,
        height = 4.5cm,
        cycle list = {
            {Reds-H, smooth, mark=none}, 
            {Blues-H, smooth, mark=none}, 
            {Reds-H, only marks, mark=*}, 
            {Blues-H, only marks, mark=*}, 
        },
    }
    \begin{axis}[
        axis y line* = left,
        xlabel = {Gate voltage \(V_{gs}\) [\si{\volt}]},
        ylabel = {Drain current \(I_{ds}\) [\si{\micro\ampere}]},
        xtick = {-0.7, -0.5, -0.3, -0.1, 0.1, 0.3, 0.5, 0.7},
        ytick = {0, 10, 20, 30, 40, 50},
        ymax = 50,
    ]
    \addplot table [x=nmos_vg_tcad_sat, y=nmos_id_tcad_sat, col sep=comma] {data/device_calibration_IdVgs.csv};
    \addplot table [x=nmos_vg_tcad_lin, y=nmos_id_tcad_lin, col sep=comma] {data/device_calibration_IdVgs.csv};
    \addplot table [x=nmos_vg_exp_sat, y=nmos_id_exp_sat, col sep=comma] {data/device_calibration_IdVgs.csv};
    \addplot table [x=nmos_vg_exp_lin, y=nmos_id_exp_lin, col sep=comma] {data/device_calibration_IdVgs.csv};
    \pgfplotsset{cycle list shift = -4}
    \addplot table [x=pmos_vg_tcad_sat, y=pmos_id_tcad_sat, col sep=comma] {data/device_calibration_IdVgs.csv};
    \addplot table [x=pmos_vg_tcad_lin, y=pmos_id_tcad_lin, col sep=comma] {data/device_calibration_IdVgs.csv};
    \addplot table [x=pmos_vg_exp_sat, y=pmos_id_exp_sat, col sep=comma] {data/device_calibration_IdVgs.csv};
    \addplot table [x=pmos_vg_exp_lin, y=pmos_id_exp_lin, col sep=comma] {data/device_calibration_IdVgs.csv};
    \draw (0, -10) -- (0, 60);
    \begin{pgfonlayer}{axis descriptions}
        \node[anchor = north west] at (0, 48) {nFinFET};
        \node[anchor = north east] at (0, 48) {pFinFET};
        \draw[-latex] (-0.35, 1) node[draw, circle, anchor = west, minimum size = 3ex] {} -- (-0.7, 1) node [midway, below] {linear};
        \node[anchor = north] at (0.6, 1) {\normalsize (a)}; 
    \end{pgfonlayer}
    \end{axis}
    \begin{axis}[
        ymode = log,
        axis y line* = right,
        axis x line = none,
        ylabel = {Drain current \(I_{ds}\) [\si{\micro\ampere}]},
        ytick = {1e-4, 1e-3, 1e-2, 1e-1, 1e0, 1e1, 1e2},
        legend style = {at={(0.5, 1.03)}, anchor=south},
        legend columns = 2,
        legend transposed = true,
        legend cell align = left,
    ]
    \addplot table [x=nmos_vg_tcad_sat, y=nmos_id_tcad_sat, col sep=comma] {data/device_calibration_IdVgs.csv};
    \addplot table [x=nmos_vg_tcad_lin, y=nmos_id_tcad_lin, col sep=comma] {data/device_calibration_IdVgs.csv};
    \addplot table [x=nmos_vg_exp_sat, y=nmos_id_exp_sat, col sep=comma] {data/device_calibration_IdVgs.csv};
    \addplot table [x=nmos_vg_exp_lin, y=nmos_id_exp_lin, col sep=comma] {data/device_calibration_IdVgs.csv};
    \pgfplotsset{cycle list shift = -4}
    \addplot table [x=pmos_vg_tcad_sat, y=pmos_id_tcad_sat, col sep=comma] {data/device_calibration_IdVgs.csv};
    \addplot table [x=pmos_vg_tcad_lin, y=pmos_id_tcad_lin, col sep=comma] {data/device_calibration_IdVgs.csv};
    \addplot table [x=pmos_vg_exp_sat, y=pmos_id_exp_sat, col sep=comma] {data/device_calibration_IdVgs.csv};
    \addplot table [x=pmos_vg_exp_lin, y=pmos_id_exp_lin, col sep=comma] {data/device_calibration_IdVgs.csv};
    \addlegendentry{SPICE \(\lvert V_{ds} \rvert\) = \SI{0.7}{\volt}}
    \addlegendentry{SPICE \(\lvert V_{ds} \rvert\) = \SI{0.05}{\volt}}
    \addlegendentry{Intel \(\lvert V_{ds} \rvert\) = \SI{0.7}{\volt}}
    \addlegendentry{Intel \(\lvert V_{ds} \rvert\) = \SI{0.05}{\volt}}
    \begin{pgfonlayer}{axis descriptions}
        \draw[-latex] (0.275, 22e-2) node[draw, circle, anchor = east, minimum size = 3ex] {} -- (0.65, 22e-2) node [midway, above] {log};
    \end{pgfonlayer}
    \end{axis}
\end{tikzpicture}%

%% file: plots/device_calibration_IdVds.tex
\begin{tikzpicture}
    \pgfplotsset{set layers = standard}
    \begin{axis}[
        scale only axis,
        width = 0.65\columnwidth,
        height = 4.63cm,
        xlabel = {Drain voltage \(V_{ds}\) [\si{\volt}]},
        ylabel = {Drain current \(I_{ds}\) [\si{\micro\ampere}]},
        xtick = {-0.7, -0.5, -0.3, -0.1, 0.1, 0.3, 0.5, 0.7},
        ytick = {0, 10, 20, 30, 40, 50},
        ymax = 50,
        legend pos = outer north east,
        cycle list = {
            {RdPu-L, smooth, mark=none}, 
            {RdPu-J, smooth, mark=none}, 
            {RdPu-H, smooth, mark=none}, 
            {RdPu-G, smooth, mark=none}, 
            {RdPu-F, smooth, mark=none}, 
            {RdPu-L, only marks, mark=*}, 
            {RdPu-J, only marks, mark=*}, 
            {RdPu-H, only marks, mark=*}, 
            {RdPu-G, only marks, mark=*}, 
            {RdPu-F, only marks, mark=*}, 
        },
    ]
    \addlegendimage{empty legend} 
    \addplot table [x=tcad_vd, y=tcad_id_0.7v, col sep=comma] {data/device_calibration_IdVds_nmos.csv};
    \addplot table [x=tcad_vd, y=tcad_id_0.6v, col sep=comma] {data/device_calibration_IdVds_nmos.csv};
    \addplot table [x=tcad_vd, y=tcad_id_0.5v, col sep=comma] {data/device_calibration_IdVds_nmos.csv};
    \addplot table [x=tcad_vd, y=tcad_id_0.4v, col sep=comma] {data/device_calibration_IdVds_nmos.csv};
    \addplot table [x=tcad_vd, y=tcad_id_0.3v, col sep=comma] {data/device_calibration_IdVds_nmos.csv};
    \addlegendimage{empty legend} 
    \addplot table [x=exp_vd_0.7v, y=exp_id_0.7v, col sep=comma] {data/device_calibration_IdVds_nmos.csv};
    \addplot table [x=exp_vd_0.6v, y=exp_id_0.6v, col sep=comma] {data/device_calibration_IdVds_nmos.csv};
    \addplot table [x=exp_vd_0.5v, y=exp_id_0.5v, col sep=comma] {data/device_calibration_IdVds_nmos.csv};
    \addplot table [x=exp_vd_0.4v, y=exp_id_0.4v, col sep=comma] {data/device_calibration_IdVds_nmos.csv};
    \addplot table [x=exp_vd_0.3v, y=exp_id_0.3v, col sep=comma] {data/device_calibration_IdVds_nmos.csv};
    \pgfplotsset{cycle list shift = -10}
    \addplot table [x=tcad_vd, y=tcad_id_0.7v, col sep=comma] {data/device_calibration_IdVds_pmos.csv};
    \addplot table [x=tcad_vd, y=tcad_id_0.6v, col sep=comma] {data/device_calibration_IdVds_pmos.csv};
    \addplot table [x=tcad_vd, y=tcad_id_0.5v, col sep=comma] {data/device_calibration_IdVds_pmos.csv};
    \addplot table [x=tcad_vd, y=tcad_id_0.4v, col sep=comma] {data/device_calibration_IdVds_pmos.csv};
    \addplot table [x=tcad_vd, y=tcad_id_0.3v, col sep=comma] {data/device_calibration_IdVds_pmos.csv};
    \addplot table [x=exp_vd_0.7v, y=exp_id_0.7v, col sep=comma] {data/device_calibration_IdVds_pmos.csv};
    \addplot table [x=exp_vd_0.6v, y=exp_id_0.6v, col sep=comma] {data/device_calibration_IdVds_pmos.csv};
    \addplot table [x=exp_vd_0.5v, y=exp_id_0.5v, col sep=comma] {data/device_calibration_IdVds_pmos.csv};
    \addplot table [x=exp_vd_0.4v, y=exp_id_0.4v, col sep=comma] {data/device_calibration_IdVds_pmos.csv};
    \addplot table [x=exp_vd_0.3v, y=exp_id_0.3v, col sep=comma] {data/device_calibration_IdVds_pmos.csv};
    \addlegendentry{\hspace{-5ex}SPICE data}
    \addlegendentry{\SI{0.7}{\volt}}
    \addlegendentry{\SI{0.6}{\volt}}
    \addlegendentry{\SI{0.5}{\volt}}
    \addlegendentry{\SI{0.4}{\volt}}
    \addlegendentry{\SI{0.3}{\volt}}
    \addlegendentry{\hspace{-5ex}Intel data}
    \addlegendentry{\SI{0.7}{\volt}}
    \addlegendentry{\SI{0.6}{\volt}}
    \addlegendentry{\SI{0.5}{\volt}}
    \addlegendentry{\SI{0.4}{\volt}}
    \addlegendentry{\SI{0.3}{\volt}}
    \draw (0, -10) -- (0, 60);
    \begin{pgfonlayer}{axis descriptions}
        \node[anchor = north west] at (0, 48) {nFinFET};
        \node[anchor = north east] at (0, 48) {pFinFET};
        \node[anchor = north] at (0.6, 1) {\normalsize (b)}; 
        \node (vgs) at (0.6, 39) {\(\lvert V_{gs} \rvert\)};
        \draw[-latex] (vgs.west) -- (0.33, 39);
        \draw[-latex] (vgs.east) -- (0.89, 42);
    \end{pgfonlayer}
    \end{axis}
\end{tikzpicture}%

%% file: plots/variability_calibration.tex
\begin{tikzpicture}
    \pgfplotsset{
        set layers = standard,
        scale only axis,
        width = 0.43\columnwidth,
        height = 3.5cm,
    }
    \begin{groupplot}[
        group style={
            group name = my plots,
            group size = 2 by 1,
            horizontal sep = 0cm,
            ylabels at = edge left,
            yticklabels at = edge left,
        },
        xlabel = {\(I_\text{on}\) [\si{\milli\ampere\per\nano\metre}]},
        ylabel = {\(I_\text{off}\) [\si{\nano\ampere\per\nano\metre}]},
        ymode = log,
        ymin = 1,
        ymax = 1000,
        ymajorgrids,
    ]
    \nextgroupplot [
        xmin = 0.7,
        xmax = 1.399,
        xtick = {0.8, 0.9, 1.0, 1.1, 1.2, 1.3},
    ]
    \begin{pgfonlayer}{axis descriptions}
        \node [anchor = west] at (rel axis cs: 0.1, 0.5) {nFinFET};
    \end{pgfonlayer}
    \nextgroupplot [
        xmin = 0.601,
        xmax = 1.3,
        xtick = {0.7, 0.8, 0.9, 1.0, 1.1, 1.2},
    ]
    \begin{pgfonlayer}{axis descriptions}
        \node [anchor = west] at (rel axis cs: 0.1, 0.5) {pFinFET};
    \end{pgfonlayer}
    \end{groupplot}
    \begin{groupplot}[
        group style={
            group name = my plots,
            group size = 2 by 1,
            horizontal sep = 0cm,
            ylabels at = edge left,
            yticklabels at = edge left,
        },
        ticks = none,
        legend pos = north west,
        legend columns = 2,
    ]
    \nextgroupplot [
        xmin = 320.6,
        xmax = 1440.669,
        ymin = 221.526,
        ymax = 1200.495,
    ]
    \addplot [thick, black] table [x index=0, y index=1, col sep=comma] {data/variability_calibration/nmos_black.csv};
    \addplot [thick, dashed, red] table [x index=0, y index=1, col sep=comma] {data/variability_calibration/nmos_red.csv};
    \addplot [only marks, mark=o, blue, opacity=0.5] table [x index=0, y index=1, col sep=comma] {data/variability_calibration/nmos_circles.csv};
    \legend{Regression line (SPICE), Regression line (Intel), Monte-Carlo simulations}
    \nextgroupplot [
        xmin = 1820.54,
        xmax = 2897.2599999999998,
        ymin = 221.526,
        ymax = 1170.495,
    ]
    \addplot [thick, black] table [x index=0, y index=1, col sep=comma] {data/variability_calibration/pmos_black.csv};
    \addplot [thick, dashed, red] table [x index=0, y index=1, col sep=comma] {data/variability_calibration/pmos_red.csv};
    \addplot [only marks, mark=o, blue, opacity=0.5] table [x index=0, y index=1, col sep=comma] {data/variability_calibration/pmos_circles.csv};
    \end{groupplot}
\end{tikzpicture}%

%% file: texfiles/Sec4_Exp.tex
\section{Evaluation and Comparisons}
\label{sec:experiments}
\begin{table*}[!t]
\centering
\caption{Properties for Graph Representation of Each Netlist
}
\label{tab:graphs}
\resizebox{0.9\textwidth}{!}{%
	\begin{tabular}{c|cccccc|cccccc|cc}
	\hline
	\textbf{Dataset} & \multicolumn{6}{c|}{\textbf{ITC-99}} & \multicolumn{6}{c|}{\textbf{EPFL}} & \multicolumn{2}{c}{\textbf{RISC-V}} \\ \hline
	\textbf{Benchmark} & \textbf{b14} & \textbf{b15} & \textbf{b20} & \textbf{b21} & \textbf{b22} & \textbf{b17} & \textbf{adder} & \textbf{max} & \textbf{bar} & \textbf{square} & \textbf{multiplier} & \textbf{divisor} &\textbf{zero-riscy} &\textbf{RI5CY}\\ \hline
	\textbf{\# Edges \purple{(\# Interconnects)}} & 9,630 & 13,627 & 21,442 & 21,268 & 32,711 & 39,774 & 3,126 & 4,530 & 5,028 & 23,739 & 52,123 & 143,261 & 35,718 & 83,652\\ \hline
	\textbf{\# Nodes \purple{(\# Gates + PIs + POs)}} & 4,569 & 5,790 & 10,090 & 9,921 & 15,236 & 16,296 & 1,846 & 2,730 & 1,748 & 13,687 & 25,265 & 72,823 & 15,032 & 35,648\\ \hline
	\textbf{Timing Constraint (ns)} & 0.46 & 0.54 & 0.53 & 0.53 & 0.51 & 0.7 & 0.42 & 1 & 0.35 & 1.3 & 3 & 25 & 1.2 & 1.2 \\ \hline
	\end{tabular}%
}
\end{table*}

\begin{figure*}[!t]
\centering
\includegraphics[width=0.85\textwidth]{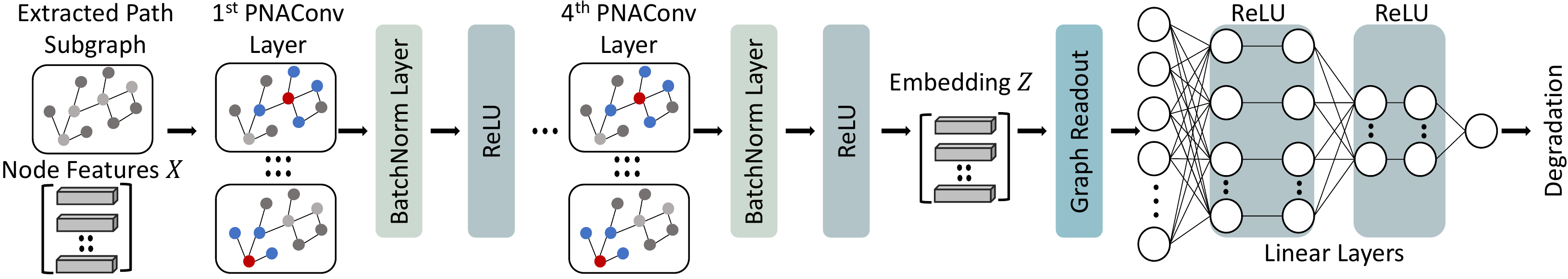}
\caption{The overall architecture of GNN4REL for reliability-degradation prediction. We utilize $4$ PNA graph convolution layers~\cite{corso2020principal}, separated by batch normalization and ReLU. A set of linear layers translates the hidden PNA output to a single value that resembles the predicted delay degradation.}
\label{fig:gnn_model}
\end{figure*}

\subsection{Experimental Setup}
\label{sec:setup}

Next, we describe the experimental setup in detail.

\subsubsection{Benchmark Designs}

We evaluate GNN4REL on selected ITC-99~\cite{itc99}
and EPFL benchmarks~\cite{amaru2015epfl}, alongside RISC-V processors~\cite{chipyard,pulp}. For the latter, we consider two different configurations of an in-order central processing unit
(CPU) as obtained using the open-source RISC-V core generator: a baseline CPU \textit{RI5CY} as well as
a lighter version \textit{zero-riscy}. RI5CY implements a $32$bit, $4$-stage CPU, while zero-riscy features
a $32$bit, $2$-stage CPU.

\subsubsection{Synthesis and Generation of Datasets}
Benchmarks are synthesized using \textit{Synopsys
Design Compiler (DC)} considering the ``fresh std-cell lib'' (i.e.,~the lib that has been characterized in the absence of variation).
Such synthesis setup provides us with {``fresh netlist''} of every benchmark. We employ std-cell libs for
the $14nm$ FinFET technology node calibrated with Intel $14nm$ FinFET measurements~\cite{bib:natarajan201414nm} (see Sec.~\ref{sec:creation}). During synthesis, highest efforts for delay minimization have been
targeted by using {``compile\_ultra''} for \textit{Synopsys DC}.
The properties for the graph representations of the synthesized gate-level netlists, alongside the
timing constraints, are listed in Table~\ref{tab:graphs}.
\textit{Synopsys PrimeTime} is used to perform STA \blue{to generate the required datasets for GNN training.} 
The aging-aware and variation-aware std-cell libs are generated as discussed in Sec.~\ref{sec:lib}.

\subsubsection{Subgraph Extraction and GNN \blue{Training}}
\label{sec:exp_gnn}
We implement the netlist-to-subgraph conversion in \textit{Perl} scripts and the subgraph extraction in \textit{Python}
scripts. The overall architecture of GNN4REL is illustrated in Fig.~\ref{fig:gnn_model}.

We use the \textit{PyTorch Geometric} implementation of PNA for graph-level regression, using four PNA GNN layers (\textit{PNAConv}) with an input/output channel size of $75$ each. A batch normalization layer
(\textit{BatchNorm}) follows each \textit{PNAConv} layer to standardize the embeddings to a mean of zero and a
variance of one. A \textit{ReLU} layer follows each \textit{BatchNorm} layer. Following the sequence of graph
convolution layers, the \textit{global\_add\_pool} \textit{Torch} function is used to generate a batch-wise
graph-level representation by averaging node features across the node dimension. The graph representation is then
passed to the following sequence of layers to obtain the final prediction: \textit{Linear}$(75, 50)$, \textit{ReLU},
\textit{Linear}$(50, 25)$, \textit{ReLU}, and \textit{Linear}$(25, 1)$. The \textit{Linear}$(input\_size, output\_size)$ layers apply a linear transformation to the incoming data. 
For the \textit{PNAConv} layers, we set the aggregators to $\{\mu, \sigma, \max, \min\}$ and the scalers to $\{identity, amplification, attenuation\}$.

To reduce the computational complexity of GNN4REL, the concept of \textit{towers} is employed in PNA as in the message
passing neural networks (MPNNs)~\cite{gilmer2017mpnn}. Using towers, the $f$-dimensional node embedding
$\bm{z}_v^{(l)}$ is broken down into $t$ different $f/t$-dimensional embeddings $\bm{z}_v^{(l,f)}$. We set the number
of towers to $5$. We train GNN4REL on the $1$-hop subgraphs for $500$ epochs using Adam optimizer with a learning rate
of $0.001$ and batch size of $32$. The model with the lowest MAE loss on the validation set is selected as the final model.

\purple{\textbf{Hyperparameters:} Since we propose a generic reliability assessment approach, we like to avoid fine-tuning/over-tuning the GNN model
	parameters such as the hidden dimensions, learning rate, etc., to obtain best performance for some given
		benchmarks. Instead, the robust results are obtained using the original parameter values of the PNA network
		proposed in~\cite{corso2020principal}. Upon the release of our GNN4REL model, readers/users may tune the
		parameters as interesting to them. Note that we study the effect of $h$ on the performance of GNN4REL in
Sec.~\ref{sec:hop}.}
		
\purple{\textbf{Fixed Architecture:} The PNA GNN shows sound performance while analyzing the circuits (represented as graphs) and automatically extracting features that are suitable for the desired tasks. Therefore, we do not need to perform any manual feature engineering or modify the network model, when considering different regression tasks (i.e., aging-induced versus process-variation-induced degradation). Rather, we only change the output information (i.e., label) and let the GNN framework perform feature engineering. We argue that this simple approach highlights the generic nature of our proposed method; the GNN4REL model can be easily trained to solve different tasks.}

\begin{figure}[!t]
\centering
\includegraphics[width=\columnwidth]{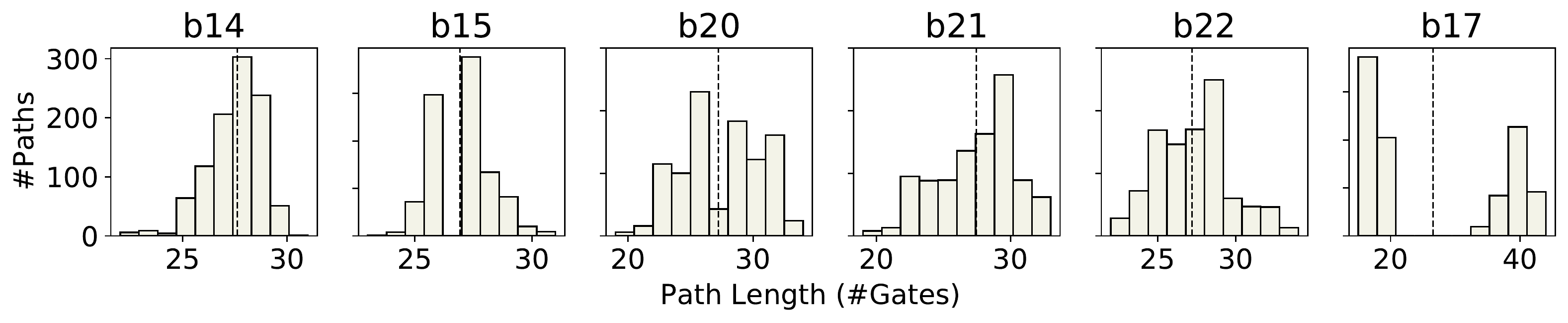}
\caption{Path-lengths distributions for the ITC-99 benchmarks. The dashed line represents the average value. }
\label{fig:itc_path_length}
\end{figure}

\begin{figure}[!t]
\centering
\includegraphics[width=\columnwidth]{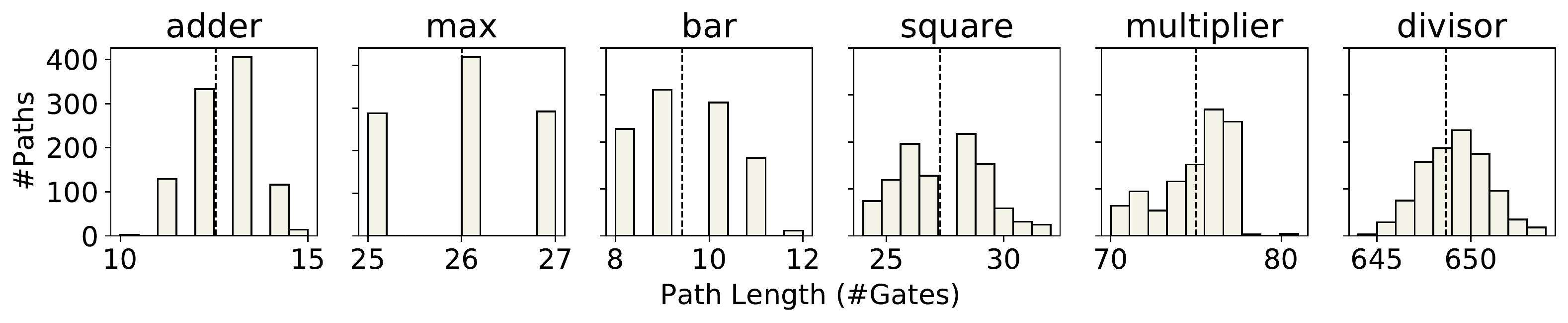}
\caption{Path-lengths distributions for the EPFL benchmarks. The dashed line represents the average value.}
\label{fig:epfl_path_length}
\end{figure}

\subsubsection{Dataset Generation and Evaluation for Prediction of Process Variation}

We consider selected ITC-99 and \purple{EPFL} benchmarks (total of \purple{$12$} netlists). Using STA, we obtain the required timings for $1,000$ extracted
timing paths from each synthesized netlist -- which form the \blue{reference} for delay-degradation computations. The
timing paths are extracted based on the $1,000$ endpoint flip-flops with the worst slack. Note that we limit the number of extracted timing paths to $1,000$ for simplicity. However, more or less timing paths can be extracted from a design based on its size and the requirements of the designer.
The
distributions of path lengths (\blue{\#gates}) for the ITC-99 \purple{and EPFL} benchmarks are shown in Fig.~\ref{fig:itc_path_length} and Fig.~\ref{fig:epfl_path_length}, respectively.

Using an in-house
\textit{Perl} script, we generate $100$ versions of each netlist by replacing each std-cell in the netlist with random and uniform sampling across the corresponding (i.e., functionally equivalent) options in the variation-aware std-cell
lib (see Sec.~\ref{sec:lib}), resulting in $1,200$ netlists and $120,000$ timing paths in total. Next, we perform STA for the same $1,000$ timing paths in the variation-affected netlists and compute the
delay-degradation percentages (i.e., the relative delay increase) by comparing the required timings after variation with the baseline timings of the paths.
The average ($\mu$), standard deviation ($\sigma$), and the maximum ($\max$) delay-degradation percentages are computed based on the STA results \blue{for a given} netlist.

\begin{figure}[!t]
\centering
\includegraphics[width=\columnwidth]{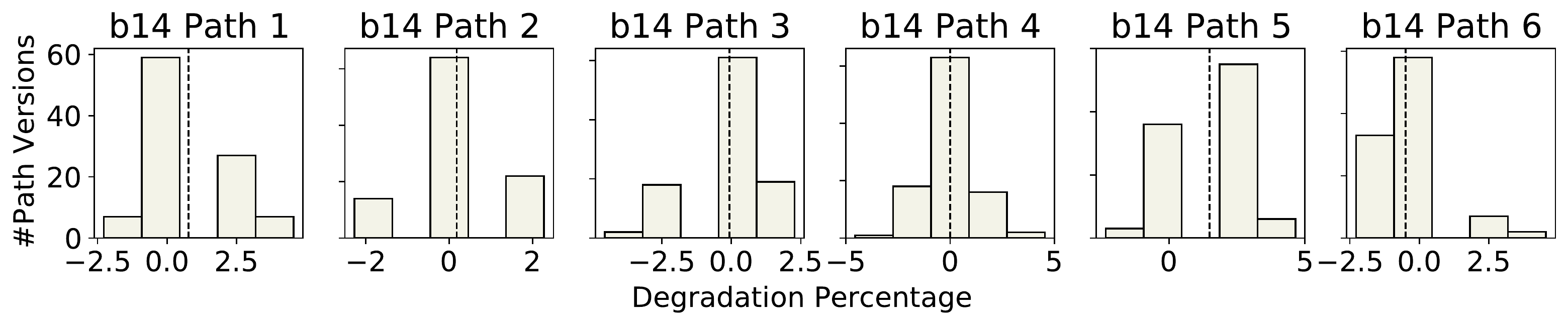}
\caption{Distributions of process-variation-induced delay degradations for selected paths of b14. The dashed line represents the average value.}
\label{fig:b14_path_delay}
\end{figure}

For an illustrative example, we randomly select six timing paths for the ITC b14 benchmark and plot the paths' delay distributions caused by process variation in Fig.~\ref{fig:b14_path_delay}. As can be observed, each path has
a unique distribution, evidencing the need for degradation prediction per path basis.

We evaluate the prediction performance of GNN4REL based on the three dataset generation scenarios discussed in
Sec.~\ref{sec:dataset_generation}. The related setup specifics are as follows:
\begin{itemize}
\item Regarding the self-referencing scenario (Sec.~\ref{sec:scenario:self-ref}), the $1,000$ timing paths of each
design are split into $810$ training paths, $100$ validation paths, and $90$ testing paths.
\item For the training on a
single-design scenario (Sec.~\ref{sec:scenario:single-des}), we consider the ITC-99 benchmarks and train GNN4REL using the b14 benchmark (without loss of generality). In this case, $900$ paths are used for training and $100$ paths are used for validation (both sets of paths extracted
		from b14).
The $1,000$ timing paths of the design under-evaluation (e.g., b15, b20, etc.) are used for testing.
In this case, the b14 benchmark is excluded from the evaluation since GNN4REL is trained on all its extracted timing paths.
\item For the design-dataset scenario (Sec.~\ref{sec:scenario:des-lib}), one of the considered ITC-99 benchmarks will be kept for testing (i.e., $1,000$
testing paths), while the rest of the considered ITC benchmarks are used for training/validation, resulting in $4,500$ training paths and $500$ validation paths.
\end{itemize}

\subsubsection{Dataset Generation and Evaluation for Prediction of Aging}

We consider the ITC-99, EPFL, and RISC-V benchmarks for this experiment. We extract $1,000$ timing paths from each
netlist along with the baseline delay values. We run STA considering the aging-aware std-cell lib for all the
benchmarks and obtain the delay-degradation percentage for each timing path. As with the case of process variation, we evaluate the prediction performance of GNN4REL based on the
three dataset generation scenarios discussed in Sec.~\ref{sec:dataset_generation}.

\begin{figure*}[!t]
\centering
\includegraphics[width=\textwidth]{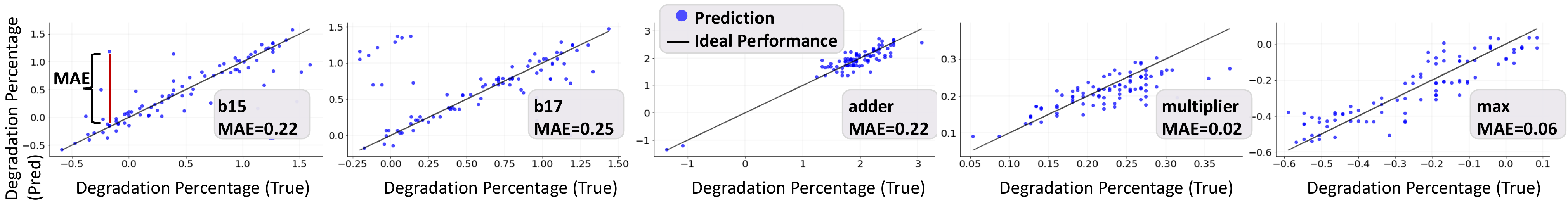} 
\caption{\purple{Path-level regression for the average ($\mu$) process variation degradation of selected ITC-99 and EPFL benchmarks under the self-referencing scenario.
\purple{Note that the ideal-performance curve is obtained using the Monte-Carlo STA~\cite{bib:klemme2021machine}.}}}
\label{fig:avg_results}
\end{figure*}
\subsubsection{Evaluation Metric}

For all experiments, the prediction performance of GNN4REL is reported using the MAE and mean absolute percentage error (MAPE) metrics, where $MAE = \frac{1}{N}\sum_{i=1}^{N}|y_i-\hat{y}_i|$ and $MAPE = \frac{1}{N}\sum_{i=1}^{N}|\frac{y_i-\hat{y}_i}{y_i}|$, where $N$ (\#testing paths) and $y$ (true degradation$\%$).

\subsection{Prediction of Process Variation}
\label{sec:process_results}
The $\mu$-degradation predictions by GNN4REL for every path per selected designs in the self-referencing scenario are shown in
Fig.~\ref{fig:avg_results}. The predictions are scatter-plotted versus the actual degradation percentages obtained
from the accurate STA results. 
A line representing the outcome of an ideal regression model is plotted to
visualize the MAE error. Here, the MAE error denotes the average distance between the prediction points to the
regression line (i.e., average absolute residual error).

GNN4REL predicts the $\mu$-degradation with an average MAE of $0.3$, where the actual degradation percentages range from $-0.5\%$ to $2\%$. In the case of the ITC-99 benchmarks, GNN4REL
performs particularly
well for the cases of b15 and b17 benchmarks, reaching an MAE of $0.22$ and $0.25$,
	 respectively, and showing a strong correlation between the predictions and the actual results. The
	 MAE values of the $\mu$-, $\sigma$-, and $\max$-degradation predictions by GNN4REL \purple{for the ITC-99 benchmarks} under all dataset generation
	 scenarios are listed in Table~\ref{tab:variability}. \purple{Further, the MAE values of the $\mu$-, $\sigma$-, and $\max$-degradation predictions for the EPFL benchmarks under the self-referencing scenario are listed in Table~\ref{tab:variability_epfl}}.

\begin{table}[!t]
\centering
 \captionsetup{justification=centering}
\caption{MAE of path-level regression for process variations on ITC-99 benchmarks, different training datasets. NA: not applicable}
\label{tab:variability}
\resizebox{0.49\textwidth}{!}{%
	\begin{tabular}{cccccccccc}
	\hline
	 & \multicolumn{3}{c}{\textbf{Self-Referencing}} & \multicolumn{3}{c}{\textbf{Training on b14}} & \multicolumn{3}{c}{\textbf{Design Dataset}} \\ \cline{2-10} 
	\multirow{-2}{*}{\textbf{Benchmark}} & \textbf{$\mu$} & \textbf{$\sigma$} & \textbf{$\max$} & \textbf{$\mu$} & \textbf{$\sigma$} & \textbf{$\max$} & \textbf{$\mu$} & \textbf{$\sigma$} & \textbf{$\max$} \\ \hline
	\textbf{b14} & 0.58 & 0.15 & 0.94 & \cellcolor[HTML]{EFEFEF} NA & \cellcolor[HTML]{EFEFEF} NA & \cellcolor[HTML]{EFEFEF} NA & 0.65 & 0.30 & 1.34 \\ \hline
	\textbf{b15} & 0.22 & 0.07 & 0.51 & 0.76 & 0.18 & 1.88 & 0.53 & 0.15 & 1.32 \\ \hline
	\textbf{b20} & 0.56 & 0.09 & 0.93 & 0.71 & 0.29 & 1.36 & 0.61 & 0.14 & 1.13 \\ \hline
	\textbf{b21} & 0.46 & 0.07 & 0.67 & 0.75 & 0.26 & 1.32 & 0.57 & 0.14 & 1.13 \\ \hline
	\textbf{b22} & 0.50 & 0.07 & 0.59 & 0.85 & 0.19 & 1.60 & 0.63 & 0.15 & 1.02 \\ \hline
	\textbf{b17} & 0.25 & 0.05 & 0.35 & 0.72 & 0.96 & 1.97 & 0.43 & 0.97 & 2.13 \\ \hline
	\end{tabular}%
}
\end{table}
\begin{table}[!t]
\centering
 \captionsetup{justification=centering}
\caption{\purple{MAE of path-level regression for process variations on EPFL benchmarks. Self-referencing scenario}}
\label{tab:variability_epfl}
\resizebox{0.49\textwidth}{!}{%
{\color{black}	\begin{tabular}{ccccccc}
	\hline
	\textbf{Benchmark} & \textbf{adder} & \textbf{multiplier} & \textbf{square} & \textbf{bar} & \textbf{max} & \textbf{divisor} \\ \hline
	\textbf{$\mu$} & 0.22 & 0.02 & 0.23 & 0.35 & 0.06 & 0.15\\ \hline
	\textbf{$\sigma$} & 0.07 & 0.01 & 0.05 & 0.08 & 0.04 & 0.04 \\ \hline
	\textbf{$\max$} & 0.59 & 0.13 & 0.55 & 0.23 & 0.19& 0.33\\ \hline
	\end{tabular}
	}
}
\end{table}

Considering \purple{the ITC-99 benchmarks} and the self-referencing scenario, GNN4REL achieves excellent
prediction performance concerning $\sigma$-degradation, with an average MAE of $0.08$, where the actual $\sigma$ values
range between $0.92$ to $3.05$ (average MAPE error of $4\%$). GNN4REL predicts the $\max$-degradation
with an average MAE of $0.66$, where the actual $\max$ values range between $1.92$ and $9.09$ (average MAPE error of $12\%$).

Regarding the single-design scenario (b14 benchmark in this study, without loss of generality), 
GNN4REL achieves an average MAE of $0.75$,
$0.37$, and $1.62$ when predicting the $\mu$, $\sigma$, and $\max$ values, respectively. This experiment shows that,
even if the GNN does not utilize the under-evaluation design during training, it can make valuable
predictions. Such capabilities are essential in practice, where training data for the design under-evaluation may not be available.
However, considering only a single design for training does limit the type of samples the model sees and
can lead to high variance.

Therefore, we further evaluate GNN4REL when using a design dataset for training; such
approach is equally valid for scenarios where training data for the design under-evaluation itself are not available. For this case (training on a design dataset), GNN4REL achieves an average MAE of $0.56$, $0.3$, and $1.34$ for predicting the $\mu$, $\sigma$, and
$\max$ values, respectively. The results show that this scenario achieves better
prediction performance compared with the single-design scenario. In short, in case where training data for the design-under-evaluation are unavailable, the designer/foundry
can (i) utilize a generic design dataset for training or (ii) track the properties of the timing paths in the testing set and
accordingly train the model on similar paths, to enhance the prediction performance as needed.

\subsection{Aging Prediction}
\begin{figure*}[!t]
\centering
\includegraphics[width=\textwidth]{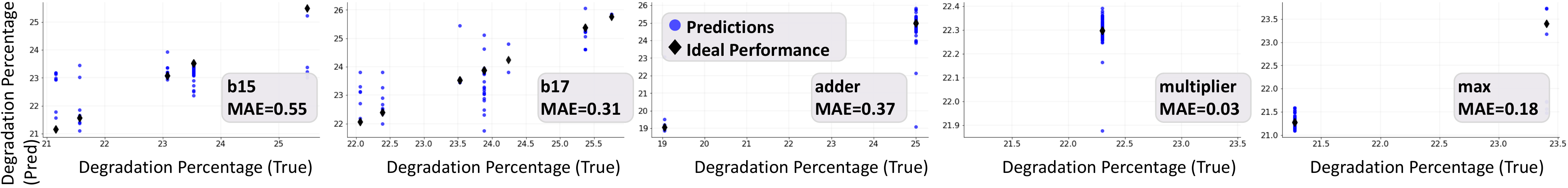}
\caption{\purple{Path-level regression for aging-induced degradation of selected ITC-99 and EPFL benchmarks under the self-referencing scenario.
\purple{Note that the ideal-performance data points are obtained using the Monte-Carlo
	STA~\cite{bib:klemme2021machine}.}}}
\label{fig:aging_results}
\end{figure*}

\begin{table}[!t]
\centering
 \captionsetup{justification=centering}
\caption{MAE path-level regression results for aging-induced degradation on selected ITC-99 benchmarks, different training datasets. NA means not applicable}
\label{tab:aging_results}
\resizebox{0.45\textwidth}{!}{%
	\begin{tabular}{cccc}
	\hline
	\textbf{Benchmark} & \textbf{Self-Referencing} & \textbf{Training on b14} & \textbf{Design Dataset} \\ \hline
	\textbf{b14} & 0.39 & \cellcolor[HTML]{EFEFEF} NA & 2.63 \\ \hline
	\textbf{b15} & 0.55 & 3.38 & 1.46 \\ \hline
	\textbf{b20} & 1.12 & 3.52 & 1.62 \\ \hline
	\textbf{b21} & 1.15 & 3.52 & 1.47 \\ \hline
	\textbf{b22} & 0.39 & 1.77 & 1.48 \\ \hline
	\textbf{b17} & 0.31 & 8.10 & 4.29 \\ \hline
	\end{tabular}%
}
\end{table}

\begin{table}[!t]
\centering
 \captionsetup{justification=centering}
\caption{MAE path-level regression results for aging-induced degradation of EPFL benchmarks. Self-referencing scenario}
\label{tab:aging_results_epfl}
\resizebox{0.49\textwidth}{!}{%
	\begin{tabular}{ccccccc}
	\hline
	\textbf{Benchmark} & \textbf{adder} & \textbf{multiplier} & \textbf{square} & \textbf{bar} & \textbf{max} & \textbf{divisor} \\ \hline
	\textbf{Self-Referencing} & 0.37 & \purple{0.03} & 0.51 & \purple{1.71} & 0.18 & 0.61\\ \hline
	\end{tabular}%
}
\end{table}

The predictions for runtime-variation degradation for every path and selected designs in the self-referencing scenario are
shown in Fig.~\ref{fig:aging_results}. The predictions are again scatter-plotted versus the actual degradation
percentages. The outcomes of an ideal regression model are also scatter-plotted, to visualize the MAE error. GNN4REL
predicts the runtime-variation degradation percentages with an average MAE of $0.65$, where the actual degradation
percentages fall between $15\%$ and $26\%$. The average MAPE value reported by GNN4REL is $3.17\%$, which indicates
an excellent prediction performance.

The MAE values for the runtime-variation degradations predicted by GNN4REL under all dataset scenarios are listed in
Table~\ref{tab:aging_results}. We can observe the same trend as with the process variation prediction: GNN4REL
performs best in case of the self-referencing scenario (average MAE of $0.65$) and worst in the single-design
dataset scenario (average MAE of $4$). Still, even in this relative worst-case scenario, GNN4REL achieves an average MAPE value of
$20\%$, which indicates that it is performing ``good forecasting'' according to the study in~\cite{lewis1982industrial}.

For the design-dataset scenario, which ``sits'' between the other two scenarios in terms of prediction performance,
GNN4REL achieves an average MAE of $2.15$ and an average MAPE of $9.7\%$, indicating on highly accurate forecasting.

We further evaluate GNN4REL under the self-referencing scenario on the EPFL benchmarks and report the MAE results in
Table~\ref{tab:aging_results_epfl}. GNN4REL achieves an average MAE of $0.64$. This experiment shows that our platform
can be generally applied to different types of designs.

For the RISC-V processors, we extract $1,000$ timing paths from the RI5CY processor and train GNN4REL accordingly.
Then, $1,000$ timing paths are extracted from the zero-riscy processor to estimate their delay degradation.\footnote{Thus, we
consider the design-dataset scenario here, but specifically for RISC-V designs with a model trained separately from those used for the earlier experiments on ITC and EPFL benchmarks.} GNN4REL reports an MAE of
$3$, demonstrating the scalability of the platform also for such more complex designs.
\subsection{\purple{Scalability Analysis}}
\label{sec:scalability}
\purple{We demonstrated the scalability of GNN4REL on complex designs such as RISC-V processors represented using graphs with up to 35,648 nodes and 3,652 edges. Moreover, we considered EPFL benchmarks represented using graphs with up to 72,823 nodes and 143,261 edges, as summarized in Table~\ref{tab:graphs}. Recall that the number of nodes represents the total number of gates, PIs and POs in the corresponding design. }

Increasing the size 
of the design does not undermine the performance of GNN4REL. Even on the contrary, larger designs can improve the prediction performance. 
		E.g., predicting the average process-variation-induced delay degradation on the smallest considered
		ITC-99 benchmark, b14 with 9,630 nodes, GNN4REL achieves an MAE of $0.58$, whereas when performing the same task on the largest considered ITC-99 benchmark, b17 with 39,774
		($4\times$ larger than b14), GNN4REL achieves the best performance with an MAE of $0.25$.

\purple{Further, the training time incurred by GNN4REL grows only linearly (with factor $<1$) with the size of the
	considered design. For example, training
	on b14 takes 01:42:18 (h:m:s) while training on b17 takes 02:36:42, i.e., a factor only $1.6\times$ training
	time although the circuit size increased to $4\times$. Additionally, in case longer training times are not desired, we have
		demonstrated how GNN4REL can also be trained on small designs such as b14 and perform inference on larger
		designs such as b17. In such as setup, GNN4REL reported an MAE of $0.72$, which is $2.9\times$ higher
		than the MAE observed for self-reference training.}

\subsection{Effect of the Extracted Subgraph Size $h$}
\label{sec:hop}

\begin{table}[!t]
\centering
 \captionsetup{justification=centering}
\caption{GNN4REL prediction performance in terms of MAE for different $h$-hop numbers under the self-referencing scenario}
\label{tab:hop-size}
\resizebox{0.49\textwidth}{!}{%
\begin{tabular}{ccccccc}
\hline
\multirow{2}{*}{\textbf{Benchmark}} & \multicolumn{3}{c}{\textbf{$\mu$ of Design-Time Degradation}} & \multicolumn{3}{c}{\textbf{Runtime Degradation}} \\ \cline{2-7} 
 & \textbf{h=0} & \textbf{h=1} & \textbf{h=2} & \textbf{h=0} & \textbf{h=1} & \textbf{h=2} \\ \hline
\textbf{b14} & 0.60 & 0.58 & 0.59 & 0.40 & 0.39 & 0.75 \\ \hline
\textbf{b15} & 0.37 & 0.22 & 0.44 & 0.69 & 0.55 & 0.71 \\ \hline
\textbf{b20} & 0.56 & 0.56 & 0.58 & 1.11 & 1.12 & 1.18 \\ \hline
\textbf{b21} & 0.53 & 0.46 & 0.55 & 1.13 & 1.15 & 1.18 \\ \hline
\textbf{b22} & 0.54 & 0.50 & 0.66 & 0.32 & 0.39 & 0.79 \\ \hline
\textbf{b17} & 0.21 & 0.25 & 0.35 & 0.59 & 0.31 & 0.57 \\ \hline
\textbf{Average} & 0.47 & \textbf{0.43} & 0.53 & 0.71 & \textbf{0.65} & 0.87 \\ \hline
\end{tabular}%
}
\end{table}

We study the effect of $h$-hop sampling on the performance of GNN4REL. We repeat the experiments for predicting the
$\mu$ delay degradation (due to process variation) and predicting the end-of-life delay degradation (due to aging) for varying
$h\in[0,2]$ with a step size of $1$. We consider the ITC-99
benchmarks for this experiment under the self-referencing scenario. See Table~\ref{tab:hop-size} for the MAE results. A hop size of $h=0$ indicates that only the gates
within each timing path itself are represented in the extracted subgraphs. As can be observed from
Table~\ref{tab:hop-size}, the prediction
performance of GNN4REL improves by increasing $h$ from $0$ to $1$. We argue that more information regarding the fan-in
and fan-out structures of the timing path gates get captured in the $1$-hop subgraphs allowing for a better estimate
of the delay degradation. However, moving to a hop size of $h=2$, the prediction performance 
drops. With
the increase in subgraph size, the properties of the timing path can get lost in the vast information captured by the
graph, degrading the performance of the prediction.

\subsection{Runtime Analysis}
\label{sec:runtime}

\begin{table}[!t]
\centering
\caption{\purple{Training time of the proposed GNN4REL platform}}
\label{tab:my-runtime}
\resizebox{0.49\textwidth}{!}{%
\begin{tabular}{ccccccc}
\hline
\textbf{ITC Benchmark} & \textbf{b14} & \textbf{b15} & \textbf{b20} & \textbf{b21} & \textbf{b22} & \textbf{b17} \\ \hline
\textbf{Training Time} & 01:42:18 & 01:37:37 & 01:45:59 & 01:49:33 & 01:48:19 & 02:36:42 \\ \hline
\purple{\textbf{EPFL Benchmark}} & \purple{\textbf{adder}} & \purple{\textbf{multiplier}} & \purple{\textbf{square}} & \purple{\textbf{bar}} & \purple{\textbf{max}} & \purple{\textbf{divisor}} \\ \hline
\purple{\textbf{Training Time}} & \purple{03:07:19} & \purple{19:24:36} & \purple{06:43:34} & \purple{04:31:17} & \purple{07:31:42} & \purple{120:01:05} \\ \hline
\end{tabular}%
}
\end{table}

\purple{{\textbf{Std-Cell Lib Generation:} Each cell in the std-cell lib is characterized under different input-signal slews
		and output-load capacitances settings, typically $7 \times 7$. Further, each rise and fall condition
		for every input pin is considered. Hence, characterizing the entire std-cell lib
			involves a huge number of SPICE simulations, which is a very time-consuming process. This
				challenge is exacerbated when having to repeat the process for Monte-Carlo-like
				std-cell characterization under variations. For instance, characterizing 100
				std-cell libs takes $\approx48$ hours on a modern high-capacity server
				(using one SPICE license).}}

\purple{{\textbf{STA}: Run on the 1,000 paths extracted for the largest considered design (i.e., the \textit{divisor} benchmark with $143,261$ gates, PIs and POs) takes 80 seconds. Running STA
	to compute the delay degradation (due to process variation) considering the 100 libs takes up to 2.2
		hours for each design.}}

\textbf{Training and Inference:} We report the training time of GNN4REL on the ITC-99 and \purple{EPFL benchmarks} \purple{(the details of the benchmarks are summarized in Table~\ref{tab:graphs})} for predicting the delay degradation caused by aging under the self-referencing scenario in Table~\ref{tab:my-runtime} -- GNN4REL takes merely $\approx2$ hours to train. \purple{Recall that in the self-referencing scenario, the $1,000$ timing paths of each design are split into $810$ training paths, $100$ validation paths, and $90$ testing paths.} Also recall that training GNN4REL is a one-time effort. \purple{The subgraph-extraction time is part of the total training time, which could be sped up using parallelism}. Once GNN4REL is trained, it can be used to assess the
reliability of any given design in the considered testing set. \textit{The inference stage of GNN4REL, i.e., the actual reliability prediction, takes a few seconds}. \purple{The experiments are performed on an Intel(R) Xeon(R) CPU $X5680$ with $64GB$ of RAM.}
	
\section{\purple{Related Work}}
\label{sec:related_work}
\subsection{Learning-based Delay Degradation Prediction}
\label{sec:related_prediction_work}
\purple{Recently, different machine learning (ML)-based
	 methods were developed to estimate delay degradation. However, these methods are limited in their
		 capabilities, as we showcase next.}

{\purple{S.~M.~Ebrahimipour \textit{et~al.}~\cite{ebrahimipour2020aadam} proposed an aging-aware delay model, termed \textit{Aadam}, tailored for
generic cell libs. In \textit{Aadam}, a separate feed-forward, fully-connected neural network (FFNN) is trained for each
cell in the lib, to capture the relation between a number of aging factors and the cell's delay
degradation. During both the training and inference stages, \textit{Aadam} first passes the gate-level
netlist to a logic simulator to compute the signal probabilities for each transistor inside each gate.
Then, the respective FFNN networks are invoked inside an STA tool, to infer the aging-induced delay of
the circuit. Thus, \textit{Aadam} eliminates the need for the generation of aging-aware std-cell
libs.} \purple{The main shortcomings of \textit{Aadam} compared to GNN4REL are as follows. First, in
GNN4REL, a single model is trained to predict degradation for different circuits containing
various std-cells, unlike for \textit{Aadam}, which requires training of as many networks as cells are in the
lib.} \purple{Second, GNN4REL eliminates the need for STA during inference, unlike \textit{Aadam}.
Third, \textit{Aadam} requires invoking circuit simulations during inference, unlike for GNN4REL, which
passes the netlist directly to the trained model, without requiring any simulations during
inference. Fourth, process variations are not considered in~\cite{ebrahimipour2020aadam}.}}

\purple{F.~Klemme~\textit{et~al.}~\cite{bib:klemme2021machine,klemme2020cell} proposed ML-based cell-lib characterization methods, which
	 can be invoked by STA tools, to obtain aging-induced and process-variation-induced degradation.
		 Similar to \textit{Aadam}, these methods do not eliminate the need for STA to compute the degradation. More recently, J.~Guo~\textit{et~al.}~\cite{guo2020novel} proposed an ML-based platform for predicting path-delay
	variations. Their platform requires the user to first compute the nominal
		delay for each path in the netlist and then uses this data for an input feature. Thus, similar to the other SOTA methods above, their platform also requires conventional STA methods at inference time, unlike GNN4REL.}

%% file: texfiles/Sec5_Conclusion.tex
\section{Conclusion}
\label{sec:conclusion}
We present GNN4REL, a machine learning-based generic platform for circuit-reliability assessment. GNN4REL empowers circuit designers to obtain fast and accurate estimations of the delay degradation imposed on their designs due to process variation and device aging. Further, GNN4REL takes the burden of generating variation-aware standard-cell libraries and running static timing analysis off the designer's shoulders, while protecting confidential foundry information. 

Our experimental evaluation on selected ITC-99 and EPFL benchmarks, alongside RISC-V processors, shows that given a timing path, GNN4REL accurately predicts the delay degradation
distributions' measures (i.e., mean, standard deviation, maximum)
caused by process variation and device aging -- with a mean
absolute error down to $0.01$ percentage points -- within few seconds. Considering different dataset and training scenarios, we show that GNN4REL can operate under various setups based on the requirements of the designer. All in all, we believe that GNN4REL opens up new frontiers in advancing design-time reliability assessment methods. We will release GNN4REL as open-source framework to the community.

%% file: bios.tex
\begin{IEEEbiography}[{\includegraphics[width=1in,height=1.25in,clip,keepaspectratio]{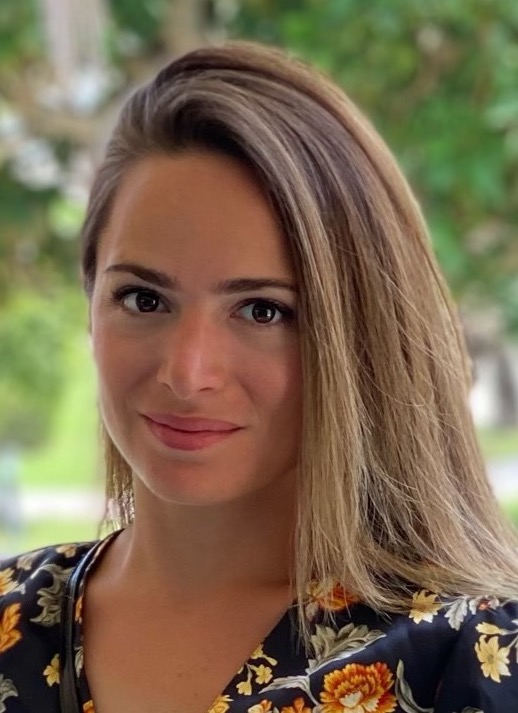}}]{Lilas~Alrahis} is a Postdoctoral Associate at New York University Abu Dhabi. She received the M.Sc.\ degree and the Ph.D.\ degree in electrical and computer engineering from Khalifa University, UAE, in 2016 and 2021, respectively. 
Her research interests include Hardware Security, Design for Trust, Logic Locking, and Applied Machine Learning. 
She won the MWSCAS Myril B.\ Reed Best Paper Award in 2016 and the Best Paper Award at the Applied Research Competition held in conjunction with Cyber Security Awareness Week, in 2019. Dr. Alrahis is currently serving as Associate Editor of the Integration, the VLSI Journal. 
\end{IEEEbiography}

\vspace{-1cm}
\begin{IEEEbiography}[{\includegraphics[width=1in,height=1.25in,clip,keepaspectratio]{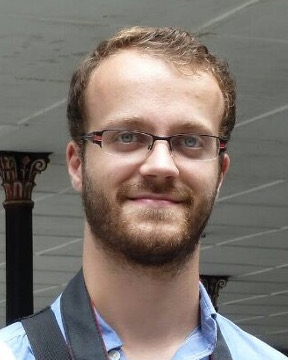}}]{Johann Knechtel}
is a Research Scientist with New York University Abu Dhabi, United Arab Emirates. 
He received the M.Sc.\ degree in Information Systems Engineering (Dipl.-Ing.) and the Ph.D.\ degree in Computer
Engineering (Dr.-Ing., summa cum laude) from TU Dresden, Germany, in 2010 and 2014, respectively. 
His research interests cover VLSI physical design automation, with particular focus on emerging technologies and
hardware security.
He was a Postdoctoral Researcher with the Masdar Institute of Science and Technology, Abu Dhabi, from 2015--2016. 
From 2010 to 2014, he was a Ph.D.\ Scholar with the DFG Graduate School on ``Nano- and Biotechnologies for Packaging of Electronic Systems'' at TU Dresden. 
In 2012, he was a Research Assistant with the Chinese University of Hong Kong, Hong Kong. 
In 2010, he was a Visiting Research Student with the University of Michigan at Ann Arbor, MI, USA. 
\end{IEEEbiography}

\vspace{-1cm}
\begin{IEEEbiography}[{\includegraphics[width=1in,height=1.25in,clip,keepaspectratio]{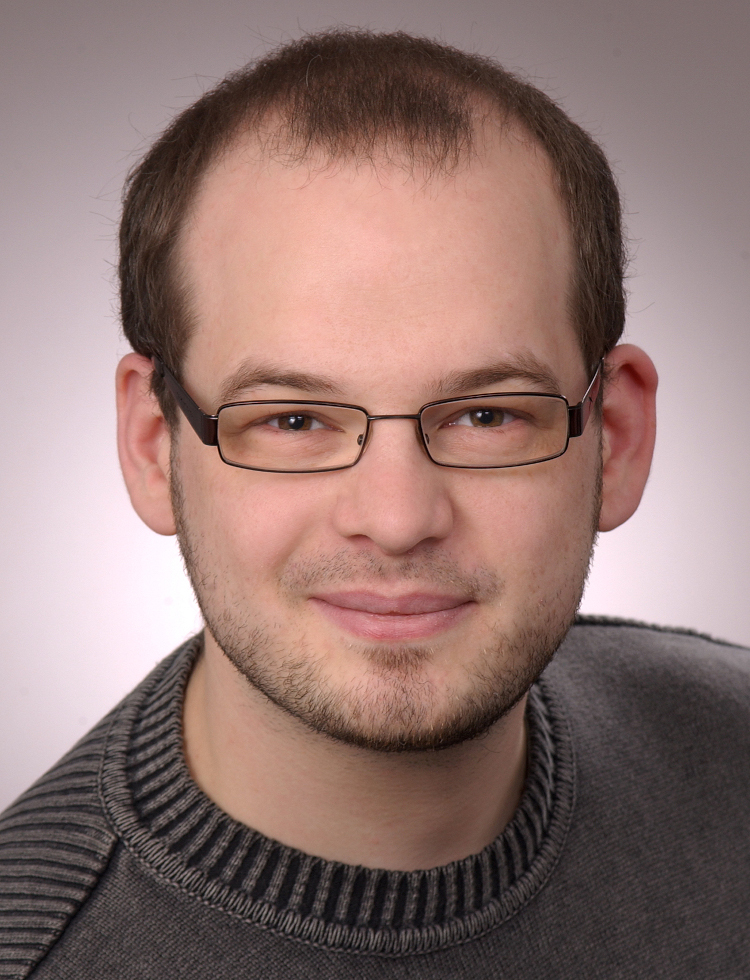}}] {Florian Klemme}(M'20) is a Doctoral Researcher at the Chair of Semiconductor Test and Reliability (STAR), University of Stuttgart.
He received the B.Sc. in System Integration from the University of Applied Sciences Bremerhaven, Germany, in 2014 and the M.Sc. in Computer Science from the Karlsruhe Institute of Technology, Germany, in 2018. He is currently working towards the Ph.D. degree at the Chair of Semiconductor Test and Reliability, University of Stuttgart. His research interests include cell library characterization and machine learning techniques in electronic design automation and computer-aided design. He is a member of the IEEE. ORCID 0000-0002-0148-0523.
\end{IEEEbiography}

\vspace{-1cm}
\begin{IEEEbiography}[{\includegraphics[width=1in,height=1.25in,clip,keepaspectratio]{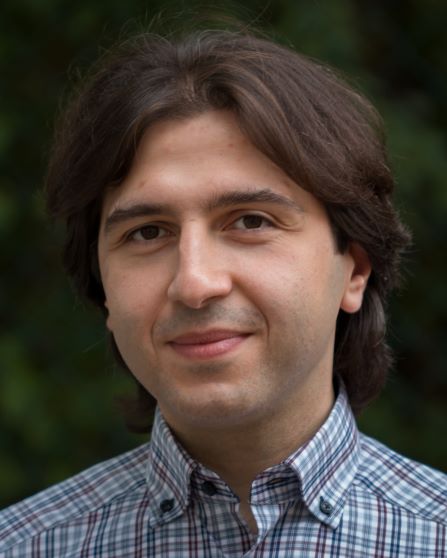}}] {Hussam Amrouch}(S'11-M'15) is a Jun.-Professor heading the Chair of Semiconductor Test and Reliability (STAR) within the Computer Science, Electrical Engineering Faculty at the University of Stuttgart as well as a Research Group Leader at the Karlsruhe Institute of Technology (KIT), Germany. He currently serves as Editor at the Nature Scientific Reports Journal. 
He received his Ph.D. degree with the highest distinction (Summa cum laude) from KIT in 2015. His main research interests are design for reliability and testing from device physics to systems, machine learning for CAD, HW security, approximate computing, and emerging technologies with a special focus on ferroelectric devices. He holds eight HiPEAC Paper Awards and three best paper nominations at top EDA conferences: DAC'16, DAC'17 and DATE'17 for his work on reliability. 
He has served in the technical program committees of many major EDA conferences such as DAC, ASP-DAC, ICCAD, etc. and as a reviewer in many top journals like Nature Electronics, T-ED, TCAS-I, TVLSI, TCAD, TC, etc. He has around $185$ publications (including $74$ journals) in multidisciplinary research areas across the entire computing stack, starting from semiconductor physics to circuit design all the way up to computer-aided design and computer architecture. His research in HW security and reliability have been funded by the German Research Foundation (DFG), Advantest Corporation, and the U.S. Office of Naval Research (ONR). 
\end{IEEEbiography}

\vspace{-1cm}
\begin{IEEEbiography}[{\includegraphics[width=1in,height=1.25in,clip,keepaspectratio]{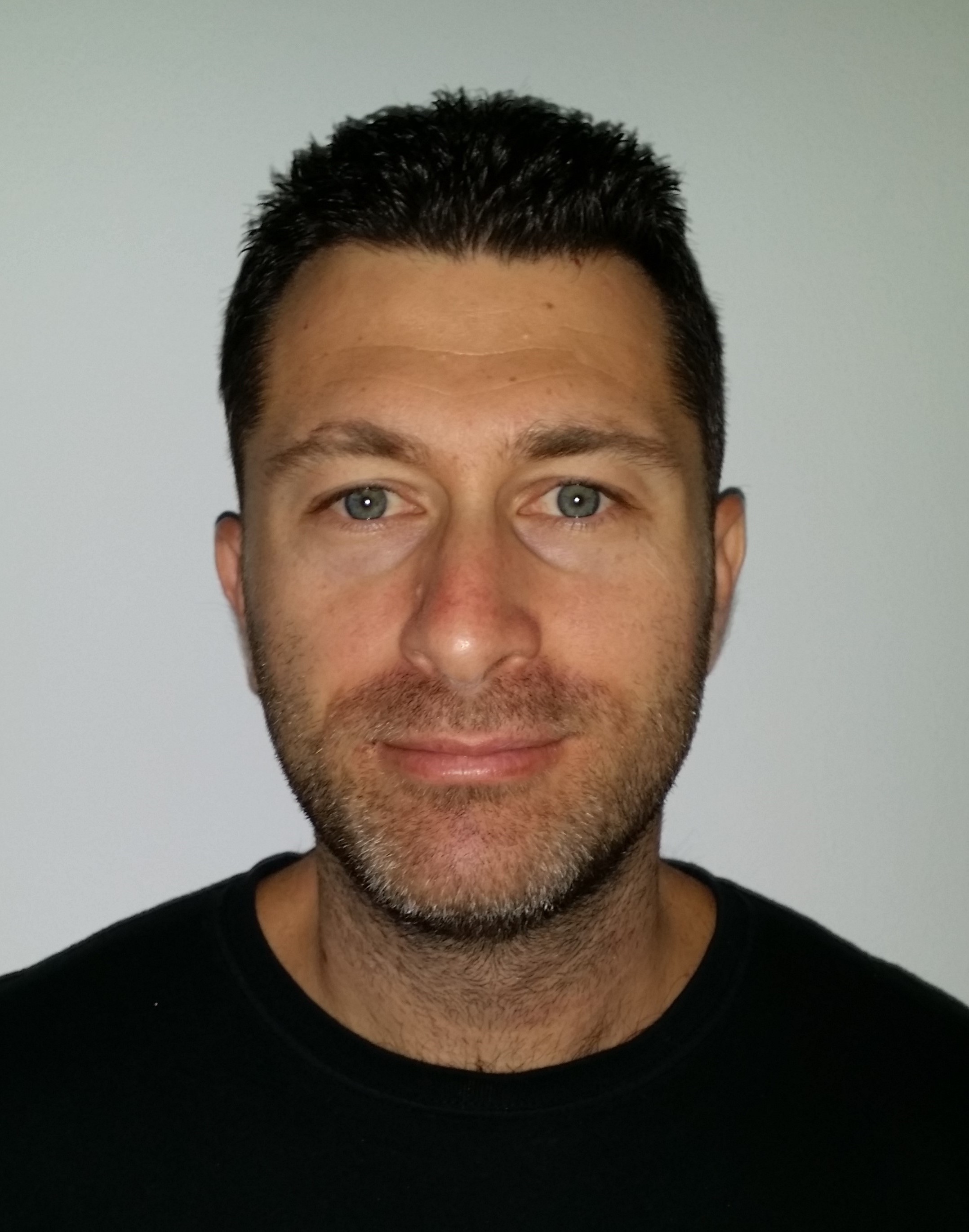}}]{Ozgur Sinanoglu} is a professor of electrical and computer engineering at New York University Abu Dhabi.
He obtained his Ph.D.\ in Computer Science and Engineering from University of California San Diego. 
He has industry experience at TI, IBM and Qualcomm, and has been with NYU Abu Dhabi since 2010. 
During his Ph.D.\ he won the IBM Ph.D.\ fellowship award twice. 
He is also the recipient of the best paper awards at IEEE VLSI Test Symposium 2011 and ACM Conference on Computer and Communication Security 2013. 
Prof.\ Sinanoglu’s research interests include design-for-test, design-for-security and design-for-trust for VLSI circuits, where he has more than 200 conference and journal papers, and 20 issued and pending US Patents.
Prof.\ Sinanoglu is the director of the Center for CyberSecurity at NYU Abu Dhabi. 
His recent research in hardware security and trust is being funded by US National Science Foundation, US Department of Defense, Semiconductor Research Corporation, Intel Corp, and Mubadala Technology.
\end{IEEEbiography}